\documentclass[lettersize,journal]{IEEEtran}
\usepackage{amsmath,amsfonts}
\usepackage{algorithm}
\usepackage{algpseudocode} 
\usepackage{array}
\usepackage[caption=false,font=normalsize,labelfont=sf,textfont=sf]{subfig}
\usepackage[dvipsnames]{xcolor}
\usepackage{textcomp}
\usepackage{makecell}
\usepackage{stfloats}
\usepackage{setspace}
\usepackage{cite}
\usepackage{algorithm}
\usepackage{algpseudocode} 
\usepackage{url}
\usepackage{verbatim}
\usepackage{graphicx}
\usepackage{caption}
\usepackage{array}
\usepackage{tabu,multirow}
\usepackage{epsfig}
\newcommand{\refFig}[1]{Fig. \ref{#1}}
\newcommand{\refTab}[1]{\ref{#1}}
\begin{document}

\title{Contour-Aware Equipotential Learning for Semantic Segmentation}

\author{Xu Yin, Dongbo Min, Yuchi Huo, Sung-Eui Yoon
\thanks{This work was supported by the National Research Foundation of Korea (NRF) grant funded by the Korea government (MSIT) (No. 2019R1A2C3002833), (No. 2021R1A4A1032582), and in part by Zhejiang Lab (121005-PI2101) (Corresponding authors: Yuchi Huo; Sung-eui Yoon.)

Xu Yin is with the School of Computing, Korea Advanced Institute of Science and Technology, Daejeon 34141, South Korea (E-mail: yinofsgvr@kaist.ac.kr).

Dongbo Min is the Faculty of the Department of Computer Science and Engineering,
Ewha Womans University, Seoul 03760, South Korea (E-mail: dbmin@ewha.ac.kr).

Yuchi Huo is with the State Key Lab of CAD and CG, Zhejiang University, China and Zhejiang Lab, China 310058 (E-mail: huo.yuchi.sc@gmail.com).

Sung-eui Yoon is with the Faculty of School of Computing, Korea Advanced
Institute of Science and Technology, Deajeon 34141, South Korea (E-mail:
sungeui@gmail.com).
}
\thanks{This paper has supplementary downloadable material available at http://ieeexplore.ieee.org., provided by the author. The material includes the source code and more experiment results. This material is 2066 kb in size.}}

\markboth{IEEE TRANSACTIONS ON MULTIMEDIA,~Vol.~1, 2022}%
{Shell \MakeLowercase{\textit{et al.}}: A Sample Article Using IEEEtran.cls for IEEE Journals}


\maketitle

\begin{abstract}
With increasing demands for high-quality semantic segmentation in the industry, hard-distinguishing semantic boundaries have posed a significant threat to existing solutions. Inspired by real-life experience, i.e., combining varied observations contributes to higher visual recognition confidence, we present the equipotential learning (EPL) method. This novel module transfers the predicted/ground-truth semantic labels to a self-defined potential domain to learn and infer decision boundaries along customized directions. The conversion to the potential domain is implemented via a lightweight differentiable anisotropic convolution without incurring any parameter overhead. Besides, the designed two loss functions, the point loss and the equipotential line loss implement anisotropic field regression and category-level contour learning, respectively, enhancing prediction consistencies in the inter/intra-class boundary areas. More importantly, EPL is agnostic to network architectures, and thus it can be plugged into most existing segmentation models. This paper is the first attempt to address the boundary segmentation problem with field regression and contour learning. Meaningful performance improvements on Pascal Voc 2012 and Cityscapes demonstrate that the proposed EPL module can benefit the off-the-shelf fully convolutional network models when recognizing semantic boundary areas. Besides, intensive comparisons and analysis show the favorable merits of EPL for distinguishing semantically-similar and irregular-shaped categories.
\end{abstract}


\section{Introduction}
\label{sec:introduction}
\IEEEPARstart{E}{xisting} deep semantic segmentation approaches \cite{FCN,unet,segnet,wsss,channel_cvpr21} are usually trained with the cross-entropy (CE) loss for multi-class classification. In the training phase, this loss measures the mismatch between the areas determined by the probability estimation from the neural network and areas defined by the ground-truth semantic label. However, the existing benchmarks' inherent drawbacks may prevent CE from achieving better performance, especially in semantic boundary areas. 

The first disadvantage relates to the label noise. It is studied \cite{noise_2,noise_3} that the misaligned ground-truth annotations with the real object edges \cite{noise_1,noise_2,noise_3} would mislead CE and results in low segmentation accuracy on the boundary regions. At the same time, the inherent inductive bias \cite{bias1, bias2} of convolutional neural networks (CNNs) is also a big barrier for CE to learn a clear semantic boundary.

\begin{figure}[t]
\begin{center}
   \includegraphics[width=8.5cm]{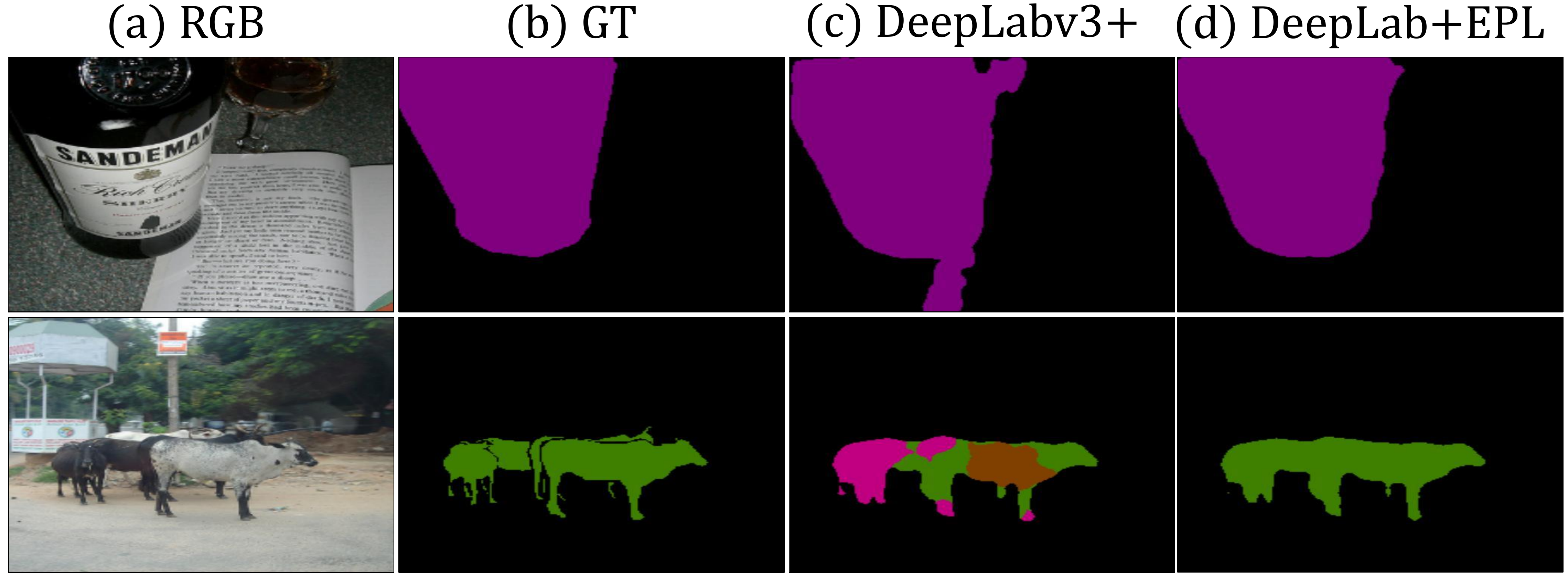}
\end{center}
   \caption{Demonstration of the semantic boundary problem. (c) 
  shows the prediction from DeepLabv3+ \cite{Deeplabv3plus}, and (d) shows the output when combined with EPL module. Both rows respectively demonstrate the segmentation difficulties in inter-class and intra-class boundary areas.}
   \label{fig:boundaryexample}
        
\end{figure}

In this work, we propose a novel method to help CE handle the boundary segmentation problem. We mainly identify the ``semantic boundary area" into two types (shown in \refFig{fig:boundaryexample}): the inter-class  and intra-class. The inter-class case refers to transition areas between different categories; these categories (on the first row) either have similar visual characteristics (patterns/textures) or have strong semantic relationships. The intra-class case (on the second row) areas are often seen when segmenting multiple instances of the same category in a small area, particularly on objects with complicated contours.

\begin{figure}[t]
\begin{center}
   \includegraphics[width=8.5cm]{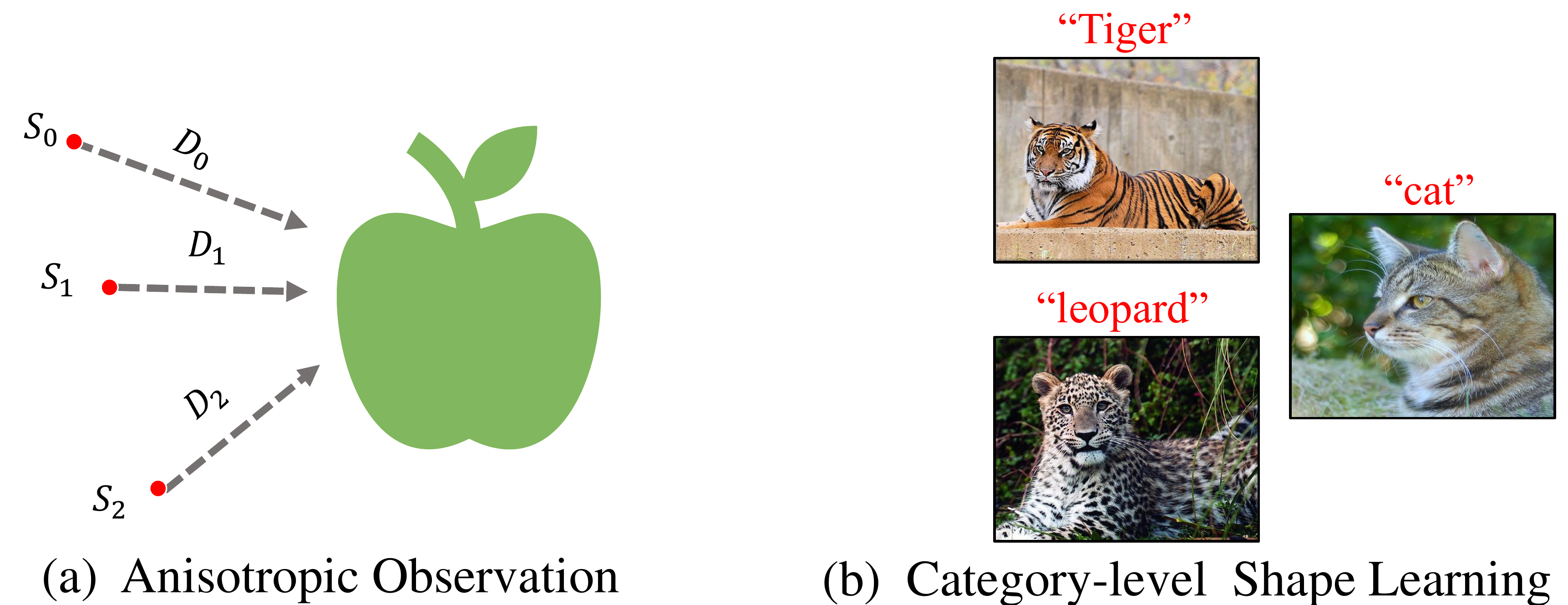}
\end{center}
   \caption{(a) This work takes inspiration from the daily visual observation, i.e., changing the relative distance $D$ and the perspective $S$  contributes the better object-contour recognition. (b) The category-level contour information is an important cue for image classification. For example, people would categorize tigers and leopards into cat species by external contours, even though they have different textures.
   }
   \label{fig:motivations}
        
\end{figure}

To this end, we present a novel framework to address the semantic boundary segmentation problem. Our central idea comes from two aspects: 
\begin{itemize}
    \item People get a good visual understanding of objects in real life by changing the relative observation distance or varying the direction/perspectives (in \refFig{fig:motivations} (a)). We conclude that better visual expression combines observations from various positions and perspectives. To this end, we propose a novel operator (anisotropic convolution) to expand the semantic labels and a loss function to refine the segmentation estimation in different directions.
    \item Objects (e.g., animals in \refFig{fig:boundaryexample}) with similar geometric appearances are usually classified as the same categories. In real life, people keep those characteristics in mind and use them as empirical evidence to recognize new species (in \refFig{fig:motivations} (b)). In this work, we suggest learning category-level contours to achieve this effect.
\end{itemize} 

Specifically, we propose the equipotential learning (EPL) module, which refines the segmentation predictions in a new domain, termed the potential domain. Following the deep segmentation tradition \cite{FCN,segnet,unet}, we define the semantic segmentation task as a pixel-wise classification problem in the conventional probability domain and then propose to convert the probability field into the potential field with anisotropic convolution to obtain visual observation from multiple perspectives. We implement two loss functions for refining the segmentation prediction in the potential field, the point loss and equipotential line loss, respectively performing anisotropic regression and category-level contour learning. 

In summary, our contributions are as follow:
\begin{itemize}
    \item We design an anisotropic convolution, a novel operator that converts the deep semantic segmentation problem to the self-defined potential domain, aiming to optimize neural network predictions from different directions. To the best of our knowledge, it is the first time to use this idea to deal with the boundary segmentation problem. 
    \item In the potential domain, we build a point loss, requiring the segmentation predictions to fit the real image content anisotropically, enforcing the consistency between predicted pixels and their nearby decision boundaries.
    \item We present an equipotential line loss to learn each category's contour. This loss specifically learns the edge regions and optimizes the corresponding predictions for better boundary segmentation performance.
    \item Experimental results on Pascal Voc 2012 and Cityscapes demonstrate that our method can help current segmentation solutions \cite{pspnet,psanet,ccnet,gscnn} refine their predictions on the semantic boundary areas.
\end{itemize}

\section{Related Work}
\label{sec:reletedwork}
\noindent\textbf{FCN methods for semantic segmentation.} FCN refers \cite{FCN} to networks adopting the convolutional layer throughout the architecture. With more similar methods \cite{unet,segnet,DeepLabv2} proposed, FCN has become the standard practice in the image segmentation community. Also, methods like conditional random field (CRF) \cite{DeepLabv2} and point-based sampling \cite{efficient,pointrend,hardpixel} are put to enforce FCN models' segmentation ability on decision boundaries. However, building specified operators brings extra computational costs. Also, these studies have relatively weak performance in learning category-level characteristics. In this work, EPL module maps each category to an independent channel,  learning category-level characteristics without increasing the parameter size.\\
\noindent\textbf{Distance field regression.}
The distance field (DF) is well-known in the computer vision and graphics community. The value of a point in the field is defined as the distance to the nearest boundary, enabling high-quality feature representation. Audebert \textit{et al.} \cite{audebert} design a multi-task model for FCN, which requires feature maps to fit a DF, apart from estimating probabilities. Recently, Xue \textit{et al.} \cite{xue} suggested networks to fit the signed distance field (SDF) directly and then use a smoothed Heaviside function to turn the distance prediction into probabilistic predictions. With incorporated information from DF, one could effectively regulate the layout of segmentation results. However, the primary issue of field-based studies~\cite{regression1,regression2,xue,IABL} is that DF itself does not carry any category identity information, which may mislead the neural network when learning multiple categories jointly. We borrow the concept of ``distance field" but map category-level image contents to independent spaces and then learn the contour information in the potential domain to address this issue. 

\noindent\textbf{Boundary supervision}: Many loss functions are specifically designed for calculating boundary loss. For instance, Discriminative Feature Network (DFN)\cite{yu} employs an edge detection step for feature maps and tries to match the edge map \cite{focalloss} by a sigmoid loss. In the medical image segmentation community, Dice loss \cite{diceloss} is widely used to solve the class-imbalance problem in the boundary region. Another loss proposed in \cite{bound_loss} re-calculates the distance field's metric in an integral regional way and achieves great success over the binary organ segmentation task. Our work uses the equipotential lines in the potential domain as the boundary supervision and learns all categories' contours with the proposed line loss.

\section{Methodology}
\label{sec:method}
This section first introduces the anisotropic convolution, an operator that convolves the semantic segmentation problem from the probability domain to the potential domain (Probability $\rightarrow$ Potential). Secondly, we elaborate on fitting anisotropic observation and performing the category-level contour learning in the potential domain. Finally, we plug EPL into FCN models to improve their boundary segmentation performance.
\subsection{Preliminaries}
Before going deep into the anisotropic convolution, we clarify two important concepts used throughout the paper.\\
\begin{itemize}
    \item \textbf{Field.} We use ``field" to represent the basic unit for domain conversion. The predicted probability fields refer to probability estimations from FCN, and the ground-truth probability fields are the one-hot encoding result of label annotations. Similarly, we name the conversion results in the potential domain, the predicted/ground-truth potential fields.
    \item \textbf{Potential energy.} We define the potential energy as the pixels' numerical value in potential fields. The anisotropic convolution converts the pixel-level probability estimations to potential energies in different directions by performing domain conversion in the training phase.
\end{itemize}
\subsection{Anisotropic Convolution for Domain Conversion}
\label{section:ac}
 To implement the domain conversion, we introduce the anisotropic convolution (AC), a differentiable convolutional operator that proceeds in specific directions. Using probability fields as the input, AC extends the image content to get its anisotropic semantic extensions. 

A general AC operator consists of a filtering kernel $W$ and anisotropic splitter $S$, corresponding to the variables of ``relative distance" and ``perspective" in the visual observation process. We let $X$ and $Y$ ($Y\in [0,1]$) stand for input images and their ground-truth probability fields. In the supervised $K$-class semantic segmentation task, we train network $f$ with the parameter $\theta$. $\hat{Y}=\{{\hat{y_{1}},\hat{y_{2}},...\hat{y_{K}}}\}$
denotes the category-level probability estimation, where:
\begin{equation}
  \hat{Y}=f_{\theta}(X),
  \label{eq:base_equation}
\end{equation}
For any point $\hat{y}^{p}\in \hat{Y}$ ($p\in P$ is the spatial coordinate set of $Y$ and $\hat{Y}$), we get its energy $E(\hat{y}^{p})$ in the potential domain by performing the  conversion on its  $w \times w$ neighborhood space $V^{\hat{p}}$. 
Formally, we express this process as:
\begin{equation}
   E(\hat{y}^{p})= V^{\hat{p}}*(W \circ S) ,\label{eq:potential_energy}
\end{equation}
where $V^{\hat{p}}$ has the same kernel size as $W$, and $*$ and $\circ$ denote the convolution and Hadamard product, respectively. Also, we apply the same conversion on all possible $y^{p}\in Y$ to get the ground-truth $E(y^{p})$.

\begin{figure}[t]
\begin{center}
\includegraphics[width=8.5cm]{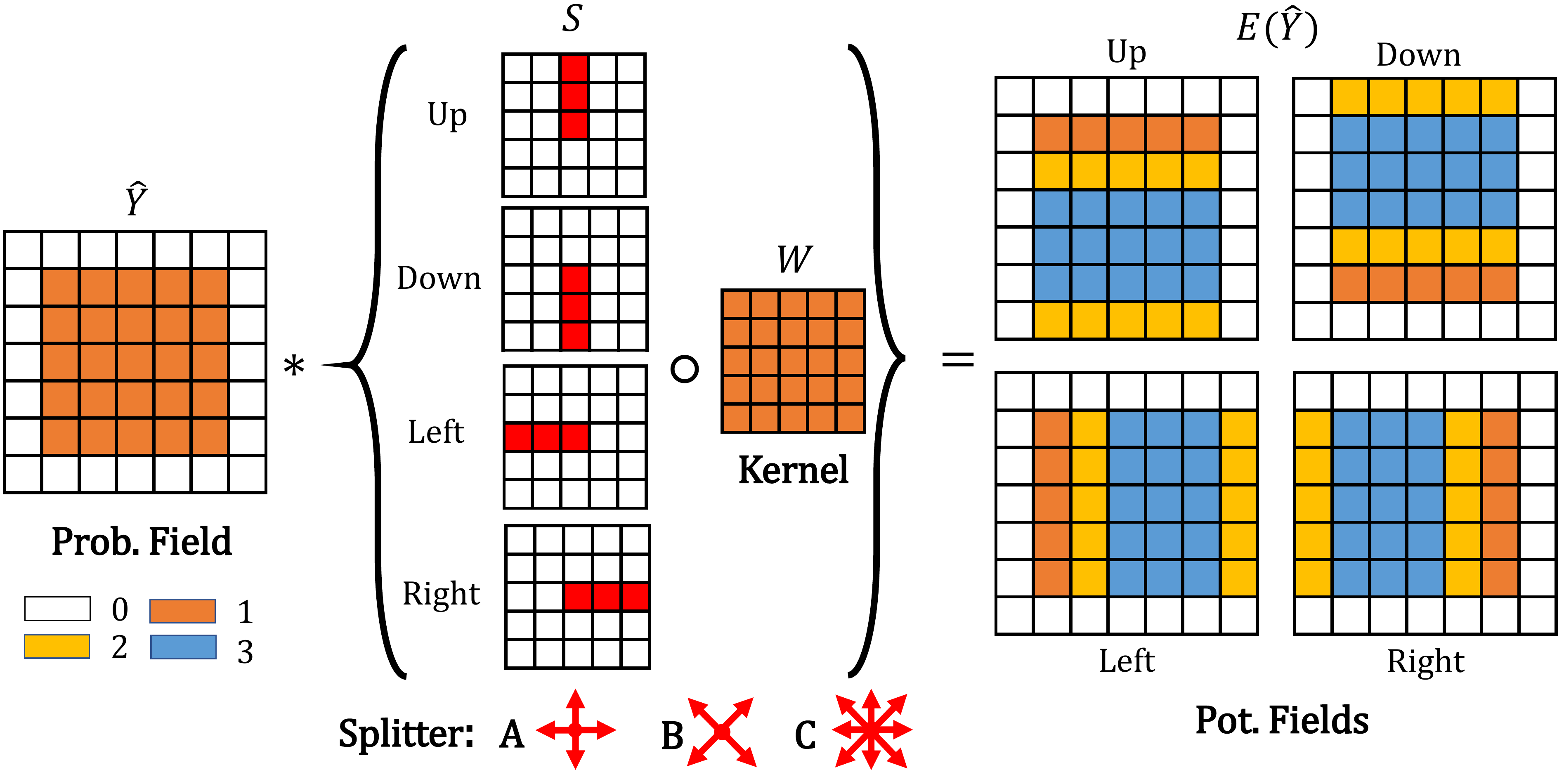}
\end{center}
   \caption{Example of the domain conversion with $5\times5$ anisotropic convolution (AC). Here, AC includes four directions (referring to splitter $A$), which expands the content contained in the probability field (\textbf{Prob. Field}) in specified directions. The potential energy of the resulting potential fields (\textbf{Pot. Fields}) ranges from 0 to 3, and we label their distribution with different colors. In experiments, we test all three splitters ($A, B$, and $C$) in ablation studies to see their effectiveness.}
   \label{fig:example_of_ac}
        
\end{figure}

In real life, we usually adjust the perspective to get different views of an object since anisotropic observations help us better understand the object. In our case, the changeable perspective is realized by letting the splitter contain different direction vectors. For instance, the splitter $S$ in \refFig{fig:example_of_ac} consists of four elements ($S=\{s_{1},s_{2},s_{3},s_{4}\}$), denoting the directions of up, down, left, and right, respectively. In the experiment section, we test three splitters $A, B$ and $C$ (shown in the bottom of \refFig{fig:example_of_ac}) to explore the effect of $S$ in semantic segmentation, containing 4, 4, and 8 directions, respectively.
To mitigate the training difficulty and reduce the computational overhead, we set the filtering kernel $W$ with a box pattern \cite{box_filter} and maintain both weights of $W$ and $S$ unchanged in the training phase. AC does not increase the parameter budget of $\theta$ in the full conversion process.

\refFig{fig:example_of_ac} presents a domain conversion example, where a $7 \times 7$ probability field $E(\hat{Y})$ is converted to four potential fields in different directions using $5 \times 5$ AC operator. We think of each potential field as an observation for the input semantic part (shown in orange) in a direction.

Generally, domain conversion has two benefits: 
\begin{itemize}
    \item \textbf{Visual:} In  \refFig{fig:example_of_ac}, we observe that the category-level semantic is extended in preset directions. We think of each potential field as a view in a specific direction. In the training phase, after converting the ground-truth probability fields to potential fields, one can get the anisotropic observations of the real image content after integrating the context information in all potential fields.
    \item \textbf{Physical:} In standard deep segmentation practices, the neural network can only get the supervision information from the semantic labels. AC spreads the image context anisotropically and maps the sparse contour label to a dense annotation in the potential fields. $E(Y)$ provides more comprehensive and stronger supervision for input $X$ than $Y$, especially in the boundary areas. Besides, domain conversion enables the network $f$ to explore a broader solution space since we can simultaneously optimize $f$ in the probability and potential domains.
\end{itemize}

Besides, AC operator enables users flexibly to change the ``observation distance" and ``perspective" variables by adjusting $W$ and $S$, based on the property of object instances (refer to Sec. \ref{sec:evluation} for details).
\subsection{Loss functions}
This section presents the point loss and the equipotential line loss that enforce the optimization in the potential domain. The first term implements the anisotropic field regression throughout the potential domain, while the second loss aims to precisely learn each category's contour.
\subsubsection*{\bf{Point Loss for anisotropic field regression}} We propose a point loss to fit the ground-truth potential fields $E(Y)$ in all directions to learn the contextual information in the potential domain. This loss regresses the fields in the potential domain globally; therefore, the information learned from the potential domain would correct prediction errors in the probability domain. 

Also, the potential energy integrates information from the neighborhood space (of the same size as $W$) for better optimization when the loss function regresses points in the semantic boundary. In other words, the computed gradient in the potential domain on $y^{p}$ will push the main segmentation network to refine $y^{p}$'s involved neighborhood predictions in the probability domain through backpropagation, enhancing the prediction consistency among pixels. 

Formally, we compute the point loss $L_{point}$ by averaging the error in each direction $s\in S$ as:
\begin{equation}
   L_{point}=\frac{1}{|S|}\displaystyle\sum_{s\in S}^{|S|}\sum_{p\in P}||E_{s}(y^{p})-E_{s}(\hat{y}^{p})||,
   \label{eq:field_loss}
\end{equation}
In the experiment section, we set the point loss with the $L1$ and $L2$ norm and then test their effectiveness. 

In conclusion, the potential fields denote the anisotropic observations for the image content; when $L_{point}$ regresses the prediction to fit the ground-truth potential fields, the segmentation network will look for an anisotropically-stable state and therefore achieve a globally-semantic balance.
\subsubsection*{\bf{Equipotential Line Loss for category-level contour learning}} Although $L_{point}$ helps improve the prediction consistency, it also brings a risk of blur boundaries because of the predicted in-between values, i.e., close to 0.5, in the potential fields (see the blur part in \refFig{fig:Abl_Point_pascal}). It would result in an intra-class indistinction problem in the category channels. Besides, the point loss designed for global regression lacks the effects of contour learning, thus failing to fully utilize the semantic boundary information in the potential domain. To solve this issue and enable contour learning, we specifically present an equipotential line loss to strengthen the optimization in the object boundary-related region.

In the ground-truth potential fields converted by the $w\times w$ AC, we define the equipotential line as a set of points having an equal energy value in the range of $[1,\lfloor \frac{w}{2}\rfloor]$. After domain conversion, the yielding ground-truth $E(Y)$ always follows a discrete distribution in $[0,\frac{w}{2}]$. By contrast, the predicted $E(\hat{Y})$'s distribution has a continuous shape since it takes the probability estimation $\hat{y}$ as input. Once matched with the ground-truth equipotential lines, the network predicts concrete category-level contours. Therefore, we suggest $E(\hat{Y})$ to specifically learn the equipotential lines that carry affluent contour information.
 
\begin{figure}[t]
\begin{center}
\includegraphics[width=8.5cm]{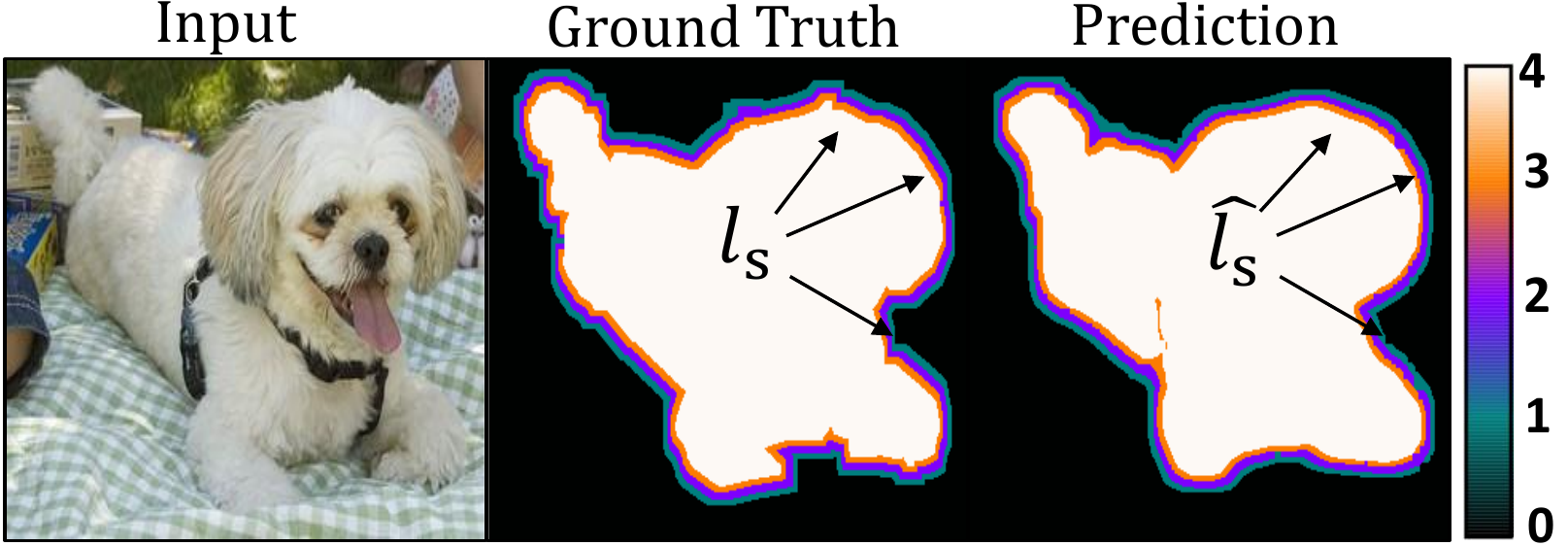}
\end{center}
   \caption{Example of the equipotential line loss for category-level contour learning with $7\times 7$ AC. We take the ``dog" category as the example and visualize its ground-truth and predicted equipotential lines (in the middle and right figure) in direction $s$, termed $l_{s}$ and $\hat{l_s}$, and mark the energy distribution with the colors in the bar. To learn the dog's contour, $L_{line}$ optimizes the mismatch region between $l_{s}$ and $\hat{l_{s}}$, in any $s \in S$, in the range [1,$\lfloor \frac{w}{2} \rfloor$] ($\lfloor \frac{w}{2} \rfloor=3$ in this example).}

   \label{fig:Line_loss_match_bin}
        
\end{figure}

Similarly, we provide an example in \refFig{fig:Line_loss_match_bin} , showing how to learn a single category's contour with $L_{line}$. With the $7\times7$ AC operating in direction $s$, we get three equipotential lines (marked with different colors). As observed in \refFig{fig:Line_loss_match_bin}, all equipotential lines are closely located around the dog's real edge and can be used to depict its contour. We generalize this example to the general $w\times w$ AC, quantify the line loss between the predicted and ground truth equipotential lines in all directions, and learn the category-level contours.

\textbf{Loss formulation:} Before elaborating on $L_{line}$'s formulation, we need to instantiate the ground-truth/predicted equipotential line regions (denoted with $L$ and $\hat{L}$) in $E(Y)$ and $E(\hat{Y})$. 

Formally, we index $E(Y)$ in the ascending order and composite $L$ by iteratively assigning $E(Y)$'s points to equipotential lines according to their energy values. When generalizing to the category-level case, for each category $i\in K$, its specified equipotential line region $l_{i,s}$ in direction $s\in S$ is represented as: $l_{i,s}=\{l_{i,s}^{1},...,l_{i,s}^{\lfloor\frac{w}{2}\rfloor}\}$; the superscript denotes the energy. 

In order to fit the continuous prediction with the discrete ground truth, we assume that each line in $\hat{l}_{i,s}$ is composed of the same number of points as its ground truth in $l_{i,s}$. Therefore, we reorder and index $E(\hat{Y})$ to let $\hat{l}_{i,s}$ be an equal-count counterpart of $l_{i,s}$. It means that for any integer $\mu \in [1,\lfloor \frac{w}{2} \rfloor]$, we have:
\begin{equation}
  |l_{i,s}^{\mu}|=|\hat{l}_{i,s}^{\mu}|,
  \label{eq:equ_count}
\end{equation}
where $|\cdot|$ denotes the number of points in the individual line.

To formulate the loss, we get inspiration from dice loss \cite{diceloss} and optimize the equipotential line loss $L_{line}$ in all directions by enlarging the intersection between each $\langle l_{i,s}$, $\hat{l}_{i,s} \rangle$ pair. 

\textbf{Algorithm~\ref{alg:line_Loss}} presents the implementation details of $L_{line}$. To compute the intersection part, we specifically apply an exponential activation with factor $\mu$ to punish large mismatches and then measure the overlapped area between $l_{i,s}$ and $\hat{l}_{i,s}$. Additionally, we use a constant value $C$ to normalize the equipotential dice coefficient (EDC) within $[0,1]$ as dice loss. It is clear to see that $L_{line}$ decreases when the corresponding lines in $l_{i,s}$ and $\hat{l}_{i,s}$ match with each other.

\begin{algorithm}[t]
\setstretch{1.1}
	\caption{Equipotential Line Loss} 	 
    \label{alg:line_Loss}
     \hspace*{\algorithmicindent} \textbf{Input:} the ground-truth/predicted line region $L,\hat{L}$, and the normalization factor $\mu$.\\
 \hspace*{\algorithmicindent} \textbf{Output:$L_{line}$} 
	\begin{algorithmic}[1]
	\State Initialized $L_{line}$ = 0;
	\For {each category $i$ in K}
	\For {each direction $s$ in S}
	\State{$l_{i,s},\hat{l}_{i,s}\leftarrow$  $L[i][s]$ , $\hat{L}[i][s]$;}\algorithmiccomment{Category-level unit}
	\For {$\tau \leftarrow 1$ to $\lfloor \frac{w}{2} \rfloor$}
	    \State {$d_{i,s} \leftarrow e^{-(l_{i,s}-\tau)^{\mu}}$};
	    \algorithmiccomment{Exponential activation}
        \State {$\hat{d}_{i,s} \leftarrow e^{-(\hat{l}_{i,s}-\tau)^{\mu}}$};
        \State {Represent the intersection area:}
        \Statex\qquad\qquad\qquad{$IoU_{i,s}\leftarrow ||d_{i,s}\cdot\hat{d}_{i,s}||_{1} $;}
        \State {Measure the mismatch area:} 
        \Statex \qquad \qquad\qquad{$C\leftarrow\frac{||d_{i,s}||_{1}}{||d_{i,s}\cdot d_{i,s}||_{1}}$;}
        \Statex \qquad \qquad \qquad {$EDC_{i,s}^{\mu}\leftarrow \frac{2\cdot C\cdot IoU_{i,s}}{||d_{i,s}||_{1}+||\hat{d}_{i,s}||_{1}}$}\algorithmiccomment{Coefficient}
        \Statex \qquad \qquad \qquad{$L_{line}\leftarrow L_{line}+(1-EDC_{s}^{i})$}
        \EndFor
        \EndFor
        \EndFor
    \State \Return {$L_{line}/|S|$}

	\end{algorithmic} 

\end{algorithm}

Intuitively, the equipotential line loss achieves the goal of contour learning by enforcing a strong geometric constraint on each category's edge area. $L_{line}$ punishes the predicted in-between values in $E(\hat{Y})$ and optimizes the predictions in semantic boundary areas from the distribution perspective. Compared with $L_{point}$, $L_{line}$ concentrates on optimizing the edge part determined by $W$, making the segmentation network better contour-representation ability. Also, $L_{line}$ integrates the line-level misalignment information of $\langle L,\hat{L} \rangle$ in different directions and, therefore, can more accurately localize the mismatch boundary, even though it happens in a small region. These two characteristics effectively help address the intra-class problem. Besides, our $L_{line}$ is more concise than other semantic boundary refinement studies \cite{xue,bound_loss} and introduces no additional parameters in the training or inference phase.
\begin{figure*}[t]
\begin{center}
\includegraphics[width=17cm]{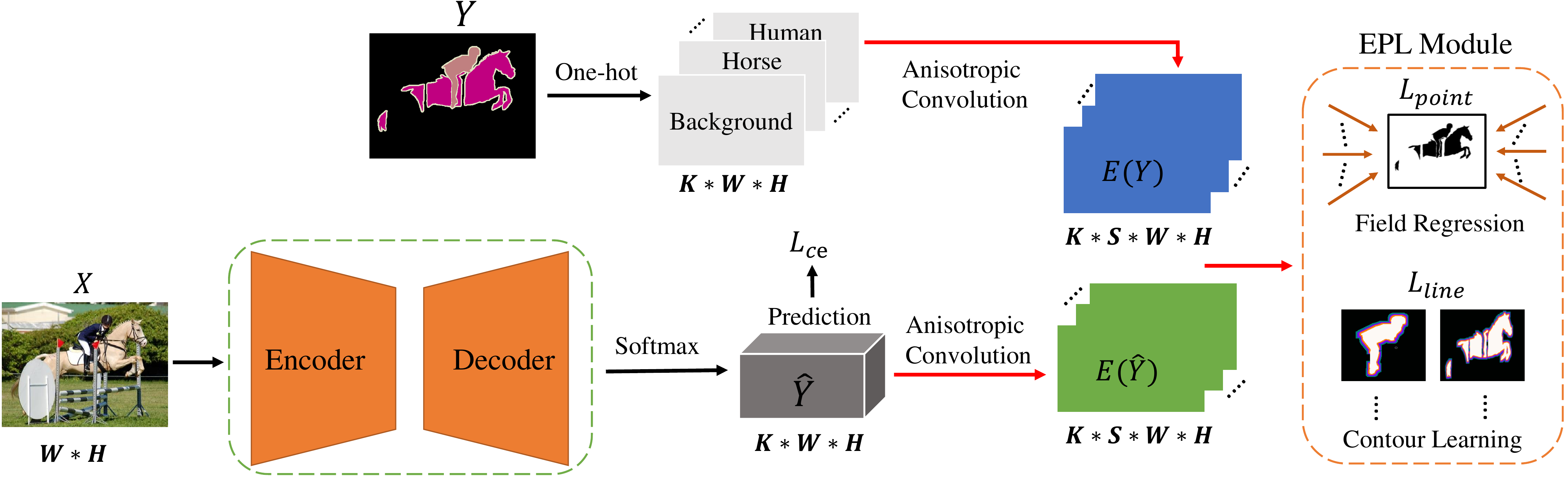}
\end{center}
   \caption{Assemble FCN model with EPL for $K$-class semantic segmentation. After proceeding with the anisotropic convolution, along any direction in $S$, the point loss $L_{point}$ and the equipotential line loss $L_{line}$ respectively enable the anisotropic field regression and the category-level contour learning.
   }
   \label{fig:example_on_fcn}
        
\end{figure*}

\subsection{Applications of EPL in FCN}
The full EPL module can be assembled to most FCN models (as illustrated in \refFig{fig:example_on_fcn}) to achieve better segmentation result. In the training phase, we follow the common practice to compute the cross-entropy loss (termed $L_{ce}$) in the probability domain. Next, we use AC to convert $Y$ and $\hat{Y}$ to the potential domain, where $L_{point}$ and $L_{line}$ are then employed for further optimization. Empirically, we use $\lambda_{1}$ and $\lambda_{2}$ as balance weights and express the final loss term as:
\begin{equation}
    Loss=L_{ce}+\lambda_{1}L_{point}+\lambda_{2}L_{line}.
    \label{eq:lossterms}
\end{equation}

In the inference stage, EPL is discarded and incurs no extra computational overhead.

\section{Evaluation}
\label{sec:evluation}
In this section, we perform two sets of experiments. In the ablation study, we add EPL module on PSPNet \cite{pspnet} to reveal the effectiveness of $L_{point}$ and $L_{line}$. Besides, we discuss the impact of anisotropic evolution (AC) when adopting different splitters. Moreover, we compare two loss functions with related studies to qualitatively evaluate their efficiency. Finally, we report the overall performance gains achieved with EPL on other baseline models.
\subsection{Setup}
\begin{itemize}
  \item \textbf{Datasets:} All experiments are conducted on two segmentation benchmarks: Pascal Voc 2012 and Cityscapes. The former dataset includes 21 classes, with 10,582 and 1,449 images for training and validation, while the latter contains 19 categories, including 2,979 training (fine-annotated) and 500 validation images.
    \item \textbf{Baseline Models:} We deploy EPL module on five FCN models: PSPNet \cite{pspnet}, PSANet \cite{psanet}, DeepLab v3+ \cite{Deeplabv3plus}, CCNet \cite{ccnet} and GSCNN \cite{gscnn}. We adopt the reliable PyTorch implementations \footnote{https://github.com/hszhao/semseg}\footnote{https://github.com/NVIDIA/semantic-segmentation}\footnote{https://github.com/jfzhang95/pytorch-deeplab-xception} to reproduce the baselines and achieve strong performances. The ablation experiments are conducted on PSPNet with resnet50 \cite{ResNet} while other baselines' performances are reported in the main result.
\item \textbf{Data augmentation and experimental details:} We strictly follow the experiment settings in the adopted implementations without changing any other hyperparameters except the loss weight $\lambda_{1}, \lambda_{2}$, and $\mu$. The involved data augmentation operations include random scale (in $[0.5, 2.0]$), random rotation (degree within $[-10, 10]$), Gaussian blur, horizontal flipping, and random crop. In the ablation part, we conduct all experiments on the small-size images with crop sizes of 256$\times$256 and 256$\times$512 on Pascal Voc 2012 \cite{pascal} and Cityscapes \cite{cityscapes} (batch=$12$). As for practical experiments, we train all models in a batch of 16 with reported image size in Table \refTab{tb:main_city} and \refTab{tb:main_pascal}. Note that all experiments are conducted on 4$\times$NVIDIA RTX Titan.
    
    \item \textbf{Evaluation protocol:} All segmentation performances are evaluated on the validation set of benchmarks. We present qualitative evaluations on each component of EPL and reveal its effect in multiple-scale inference results, in the range of [0.5, 0.75, 1.0, 1.25, 1.5, 1.75]. Except \textbf{for} the mean Intersection-over-union (mIoU), we specially employ \textbf{F-Measure} \cite{efficient, F_measure}  and \textbf{Trimap IoU} \cite{DeepLabv2,boundary_iou} to quantize the models' segmentation performance in semantic boundary areas. 
\end{itemize}

\subsection{Ablation Study}
\label{section:abl_exp}
We use \textbf{Point} and \textbf{Line} to denote the point loss and the equipotential line loss, respectively. In this part, we mainly apply the splitter A  (shown in \refFig{fig:example_of_ac}) and set it with different kernel sizes to evaluate the effectiveness of both loss functions. To make a fair comparison, we empirically set $\mu$ (Eq. \ref{eq:equ_count}), $\lambda_{1},\lambda_{2}$ (Eq. \ref{eq:lossterms}), as 10, 0.1, and 0.01, respectively. Discussions of other splitters ($B$ and $C$) and choices of hyperparameters are reported in Appendix \ref{section: append_comparison}.

\textbf{Ablation for the Point loss.} We add $L_{point}$ to the baseline network to regress the potential fields. To verify its effect, we respectively set $L_{point}$ (Eq. \ref{eq:field_loss}) of $L1$ and $L2$ norm (marked with the superscript) and report their results in Table \refTab{tab:abl_point}. One can see that we achieve considerable improvements over both datasets. We do not observe obvious performance differences between the two norms and therefore set $L_{point}$ of $L2$ norm in the later experiments. 

\begin{figure}[t]
\begin{center}
\includegraphics[width=8.5cm]{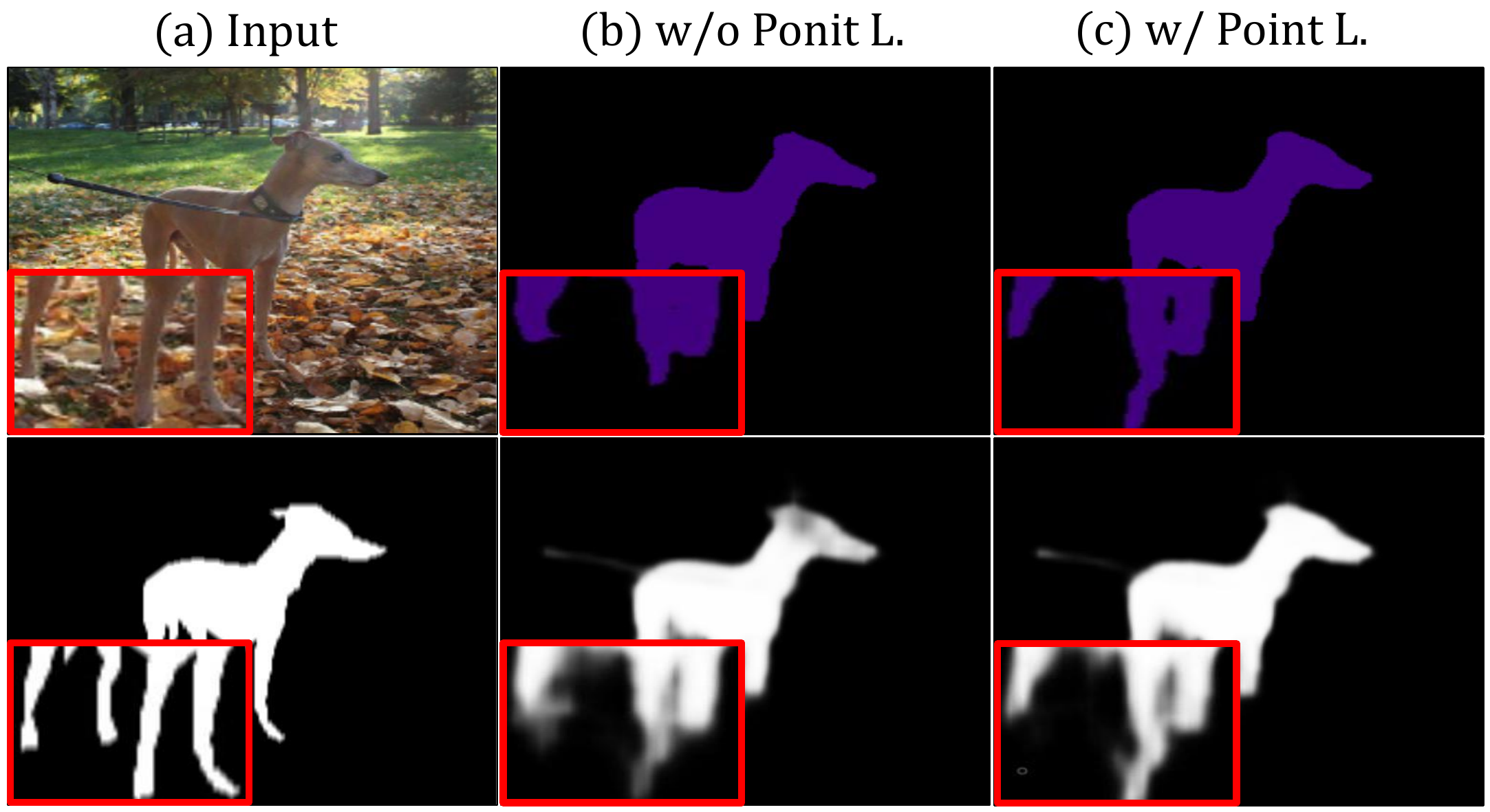}
\end{center}
   \caption{Example of the point loss on Pascal Voc 2012. In the second row, we visualize the potential field.
   }
   \label{fig:Abl_Point_pascal}
        
\end{figure}

In \refFig{fig:Abl_Point_pascal}, we see that the prediction of the dog is considerably refined by $L_{point}$. However, after visualizing one potential field (the second row), we still observe blurs between the paws, indicating the in-between predictions.

\begin{table}[t]
\small
    \centering
    \begin{tabu}{c|c<{\centering}c<{\centering}c<{\centering}}
         \textbf{Dataset}&\textbf{Method}&\textbf{Kernel}&\textbf{mIoU(\%)}\\
                                        \tabucline[0.8pt]{-}                       
         \multirow{7}{*}{Pascal Voc}&Baseline&-&73.52\\

         ~&\multirow{3}{*}{+Point$^1$}&7&74.07\\
         ~&~&9&73.85\\
         ~&~&11&\textbf{74.08\tiny{(+0.56)}}\\
            \cline{2-4}
                  ~&\multirow{3}{*}{+Point$^2$}&7&74.09\\
         ~&~&9&\textbf{74.42\tiny{(+0.90)}}\\
         ~&~&11&74.27\\
         \hline
                  \multirow{7}{*}{Cityscapes}&Baseline&-&71.66\\

         ~&\multirow{3}{*}{+Point$^1$}&9&70.44\\
         ~&~&11&70.96\\
         ~&~&13&\textbf{72.71\tiny{(+1.05)}}\\
         \cline{2-4}
                  ~&\multirow{3}{*}{+Point$^2$}&9&\textbf{72.39\tiny{(+0.73)}}\\
         ~&~&11&70.93\\
         ~&~&13&71.05\\
         \hline

    \end{tabu}
    \caption{\textbf{Point loss} performance on Pascal Voc 2012 and Cityscapes validation set. The superscript of “+Point” denotes the norm type.}
    \label{tab:abl_point}
\end{table}

\begin{figure}[t]
\begin{center}
\includegraphics[width=8.5cm]{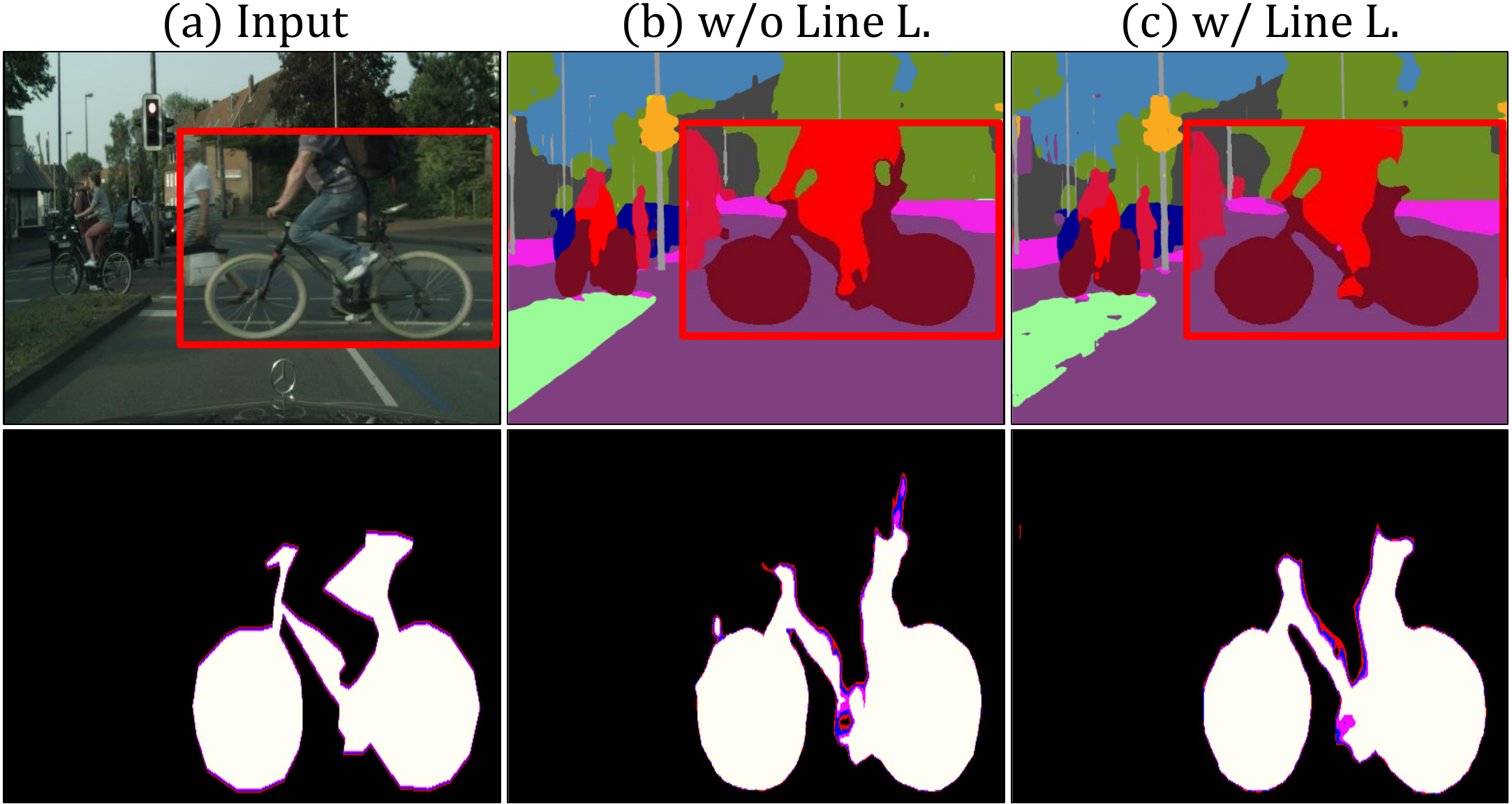}
\end{center}
   \caption{
   Example results of the equipotential line loss on Cityscapes. In the second row, we compare the visualized potential fields of the red cropped bicycle area.
 }
   \label{fig:Abl_line_pascal}
     
\end{figure}

\textbf{Ablation for the equipotential line loss.} Table \refTab{tab:abl_line} reports the performance of $L_{line}$ on both datasets. After learning the category-specific contours, we observe that PSPNet achieves up to 0.63\%/1.60\% mIoU increasements on Pascal Voc 2012/Cityscapes. Similarly, we visualize the contour learning effects in \refFig{fig:Abl_line_pascal} and observe that $L_{line}$ learned the subtle and complicated shape features of the bicycle category.

\begin{figure}[t]
\begin{center}
\includegraphics[width=8.5cm]{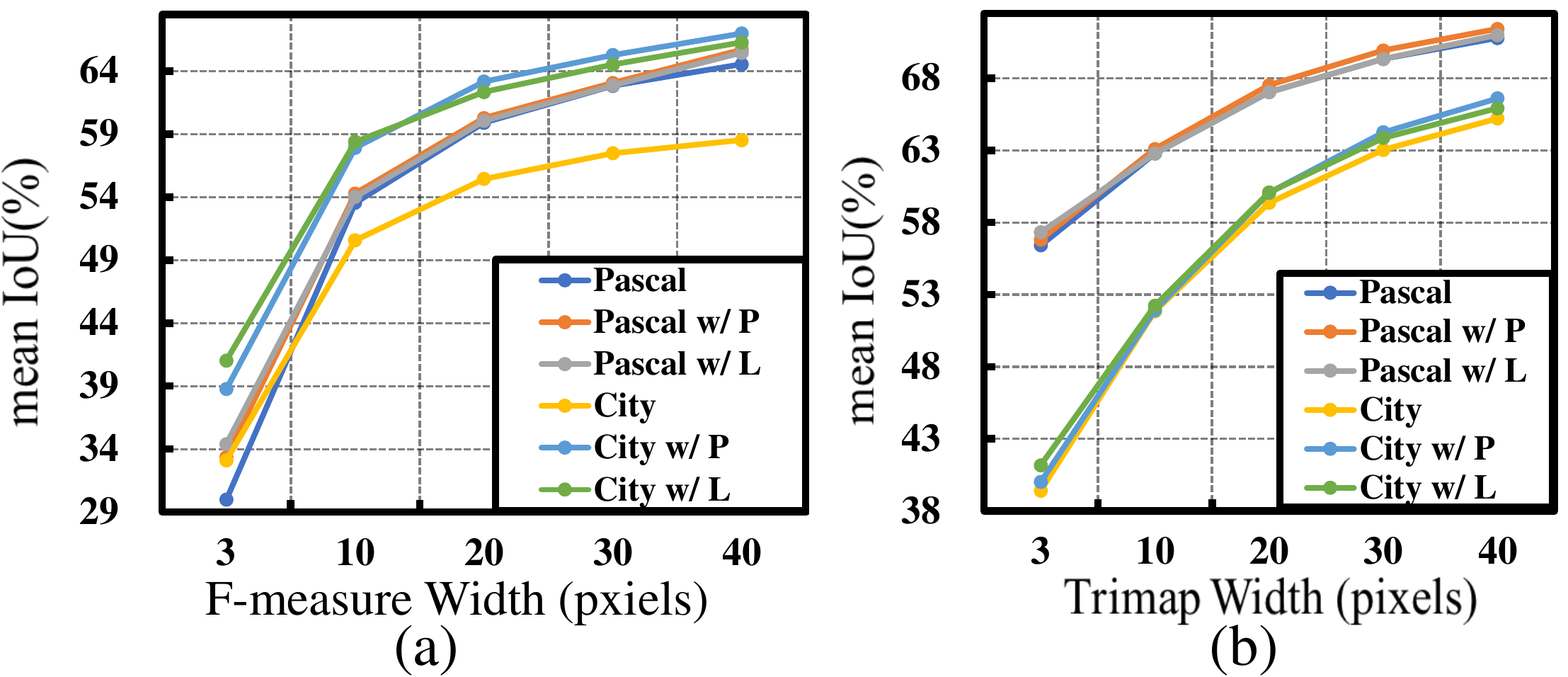}
\end{center}
  \caption{(a)(b) plot how F-measure and Trimap mIoU change with the pixel width of the ground-truth boundary region on Pascal Voc 2012 (\textbf{Pascal}) and Cityscapes (\textbf{City}) before and after employing $L_{line}$  (\textbf{w/L.}) and  (\textbf{w/P.}).
   }
   \label{fig:boundary_evaluation}
\end{figure}

\textbf{Boundary Segmentation evaluation:} To furtherly verify the learning effect of both losses, we introduce two boundary-quality measures, F-measure \cite{F_measure}, and Trimap \cite{efficient, DeepLabv2}, to evaluate segmentation results in the semantic boundary area. Both metrics measure the matching level between the prediction and the ground truth in a narrow band region from the ground-truth semantic boundary given a pixel width. We evaluate the segmentation results multiple times with different pixel widths and present the comparison with/without employing $L_{line}$ in \refFig{fig:boundary_evaluation}; one can see that both $L_{point}$ and $L_{line}$ improve the segmentation capability in the boundary areas. Besides, we see that $L_{line}$ achieves more performance enhancements than $L_{point}$ on the areas near the edge (pixel width $<10$), indicating $L_{line}$ 's effectiveness on contour learning.

\begin{table}[t]
\small
    \centering
    \begin{tabu}{c|c<{\centering}c<{\centering}c<{\centering}}
         \textbf{Dataset}&\textbf{Method}&\textbf{Kernel}&\textbf{mIoU(\%)}\\
                                        \tabucline[0.8pt]{-}  
         \multirow{4}{*}{Pascal Voc}&Baseline&-&73.52\\

         ~&\multirow{3}{*}{+Line}&7&\textbf{74.15\tiny{(+0.63)}}\\
         ~&~&9&73.47\\
         ~&~&11&74.05\\
         \hline
                  \multirow{4}{*}{Cityscapes}&Baseline&-&71.66\\

         ~&\multirow{3}{*}{+Line}&9&71.78\\
         ~&~&11&\textbf{73.26\tiny{(+1.60)}}\\
         ~&~&13&72.11\\
         \hline

    \end{tabu}
    \caption{\textbf{The equipotential line loss} performance on Pascal Voc 2012 and Cityscapes validation set.}
    \label{tab:abl_line}
\end{table}

\textbf{Effects of anisotropic convolution:} In this part, we ablate the AC operator with the standard convolution (\textbf{SC}) to verify its efficiency. Empirically,  SC  destroys the ground-truth image content and distorts its probability distributions. In Table \refTab{tb:comparison_sc}, we replace AC with SC and optimize the potential fields with $L_{point}$ and $L_{line}$ similarly. We observe that SC degrades the baseline performance on both datasets, confirming the advantages of AC and the feasibility of our "anisotropic observation" assumption.

\textbf{Effects of directional splitters:} The choice of the splitters has a crucial influence on EPL. We test the other two splitters $B$ and $C$ (reported in Appendix, Table \refTab{tab:appendix_point_abl} \& \refTab{tab:appendix_line_abl}) and observe that splitter $A$ performs the best among all three candidates and can achieve consistent improvements on both datasets, indicating that the best semantic observation comes from the integrated result in the up, down, right, and left directions.

\textbf{Effects of Kernel size:} In Table \refTab{tab:abl_point} \& \refTab{tab:abl_line}, we see that both losses' performances do not necessarily increase with AC's kernel size. We see $W$'s size as the range of semantic area that aims to optimize. In ablation experiments (Table \refTab{tab:abl_point}, \refTab{tab:abl_line}, \refTab{tab:appendix_point_abl} and \refTab{tab:appendix_line_abl}), we tested multiple kernels and found that the large ones do not work well with small object instances because the decision boundary area becomes negligible in these cases. Therefore, we use the optimal range of kernel sizes found in the ablation part, then scale AC in proportion to fit other network output resolutions in later experiments.

\textbf{Effects of the exponential activation:} To implement $L_{line}$, we apply an exponential action factor  $\mu$ to measure the overlapped part between the ground-truth $L$ and predicted $\hat{L}$(see line 6, 7 in Algorithm \ref{alg:line_Loss}) where $\mu$ must be even. Here, we test the effect of $\mu$ with five even values ${2,4,10,16,20}$, and then experiment $L_{line}$ ($\lambda_{1}=0, \lambda_{2}=0.2$) with PSPNet on two datasets.  Table \refTab{tab:hyperparameter_activation} presents PSPNet's performance when setting $\mu$ with different values. We observe that the best segmentation performances on Pascal Voc 2012 and Cityscapes are achieved when $\mu=10, 2$. Therefore, we apply both values in later experiments.

\begin{table}
\centering
\small
\begin{tabu}[t]{m{3cm}|m{1.5cm}<\centering|m{1.5cm}<\centering}

\textbf{Method} & \textbf{mIoU(\textbf{P.\%)}} & \textbf{mIoU(\textbf{C.\%)}} \\

                                        \tabucline[0.8pt]{-}  
PSPNet(baseline) & 73.52&71.66 \\

PSPNet+SC+$L_{point}$& 71.08\tiny{(-2.44)}&68.37\tiny{(-3.29)}\\

PSPNet+SC+$L_{point}$& 72.17\tiny{(-1.35)}&70.93\tiny{(-0.73)}\\
PSPNet+AC+$L_{line}$& \textbf{74.84\tiny{(+1.32)}}&\textbf{72.39\tiny{(+0.73)}}\\

PSPNet+AC+$L_{point}$& \textbf{74.71\tiny{(+1.19)}}&\textbf{73.26\tiny{(+1.60)}}\\

\hline

\end{tabu}
\caption{Compare results of AC with standard convolution (\textbf{SC}) on Pascal Voc 2012 (\textbf{P.}) and Cityscapes (\textbf{C.}).}
\label{tb:comparison_sc}
     
\end{table}

\begin{table}[t]
\small
    \centering
    \begin{tabu}{ccc}
         $\mu$&\textbf{mIoU(P.\%)}&\textbf{mIoU(C.\%)} \\
    \tabucline[0.8pt]{-} 
                 Baseline&73.52&71.66\\
         2&73.12&\textbf{72.37}\\
         4&73.73&71.04\\
         10&\textbf{74.71}&71.47\\
         16&73.85&65.51\\
         20&73.60&66.51\\
         \hline
    \end{tabu}
    \caption{Effects of $\mu$ on segmentation results on Pascal Voc 2012 (\textbf{P.}) and Cityscapes (\textbf{C.})}
    \label{tab:hyperparameter_activation}
         
\end{table}

\begin{table*}[t]
\begin{center}
\scriptsize
\begin{tabu}{m{0.8cm}<{\centering}|m{1.3cm}<{\centering}|m{0.5cm}<{\centering}|m{0.3cm}<{\centering}m{0.3cm}<{\centering}m{0.3cm}<{\centering}m{0.3cm}<{\centering}m{0.3cm}<{\centering}m{0.3cm}<{\centering}m{0.3cm}<{\centering}m{0.3cm}<{\centering}m{0.3cm}<{\centering}m{0.3cm}<{\centering}m{0.3cm}<{\centering}m{0.3cm}<{\centering}m{0.3cm}<{\centering}m{0.3cm}<{\centering}m{0.3cm}<{\centering}m{0.3cm}<{\centering}m{0.3cm}<{\centering}m{0.3cm}<{\centering}m{0.3cm}<{\centering}}
\textbf{Method} &\textbf{Backbone}& \textbf{mIoU}&\textbf{road}&\textbf{swalk}&\textbf{build}&\textbf{wall}&\textbf{fence}& \textbf{pole}&\textbf{tligh.}&\textbf{tsign}&\textbf{veg}&\textbf{terr.}&\textbf{sky}&\textbf{pers.}& \textbf{rider}&\textbf{car}&\textbf{truck}&\textbf{bus}&\textbf{train}&\textbf{mcyc}&\textbf{bcyc}\\
\tabucline[0.8pt]{-} 
PSANet&ResNet-101& 78.01&98.1&85.1&92.5&54.1&60.9& 64.7&70.8&78.6&92.6&65.5&94.6&82.9&63.2&95.0&74.9&88.0&77.6&65.1&78.0\\
+EPL&ResNet-101& \textbf{79.46}&96.8&83.7&92.9&\textbf{56.7}&\textbf{62.7}&\textbf{68.8}&\textbf{75.5}&\textbf{81.6}&93.3&66.0&94.7&\textbf{85.2}&\textbf{66.9}&95.6&70.3&\textbf{90.3}&76.7&\textbf{71.5}&\textbf{80.5}\\
\hline 
PSPNet&ResNet-101& 78.35&98.4&86.3&92.9&55.3&63.1& 65.9&73.1&79.8&93.0&66.0&94.9&83.8&64.8&95.1&74.0&85.8&75.8&71.1&79.3\\
+EPL&ResNet-101&\textbf{79.44}&96.7&83.5&\textbf{94.1}&\textbf{56.3}&62.4&\textbf{69.4}&73.7& \textbf{82.1}&92.7&\textbf{67.4}&94.2&\textbf{86.4}&\textbf{66.3}&95.7&72.3&\textbf{88.5}&\textbf{78.1}&\textbf{72.3}&80.2\\
\hline
DeepLab&WResNet-38& 79.38&98.2&85.7&92.7&60.1&62.9& 66.6&70.4&79.2&92.7&66.6&94.6&82.7&64.2&95.2&80.5&89.6&80.9&67.1&78.3\\
+EPL&WResNet-38& \textbf{80.34}&97.5&84.3&92.6&\textbf{61.2}&\textbf{64.1}&66.5&70.7& 80.1&90.5&\textbf{68.1}&94.7&\textbf{84.5}&65.1&94.9&\textbf{81.6}&\textbf{90.7}&\textbf{82.1}&67.4&\textbf{79.5}\\
\hline

\end{tabu}
\end{center}
\caption{Category-level mIoU Comparison on Cityscapes validation set. }
\label{tb:perclassL_city_PSPNET}
\begin{center}
\scriptsize
\begin{tabu}{m{0.8cm}<{\centering}|m{1.3cm}<{\centering}|m{0.5cm}<{\centering}|m{0.3cm}<{\centering}m{0.3cm}<{\centering}m{0.3cm}<{\centering}m{0.3cm}<{\centering}m{0.3cm}<{\centering}m{0.3cm}<{\centering}m{0.3cm}<{\centering}m{0.3cm}<{\centering}m{0.3cm}<{\centering}m{0.3cm}<{\centering}m{0.3cm}<{\centering}m{0.3cm}<{\centering}m{0.3cm}<{\centering}m{0.3cm}<{\centering}m{0.3cm}<{\centering}m{0.3cm}<{\centering}m{0.3cm}<{\centering}m{0.3cm}<{\centering}m{0.3cm}<{\centering}m{0.3cm}<{\centering}m{0.3cm}<{\centering}}
\textbf{Method} & \textbf{Backbone}&\textbf{mIoU}&\textbf{bgr}&\textbf{aero}&\textbf{bicy}&\textbf{bird}&\textbf{boat}& \textbf{bottle}&\textbf{bus}&\textbf{car}&\textbf{cat}&\textbf{chair}&\textbf{cow}&\textbf{dt.}&\textbf{dog}&\textbf{horse}&\textbf{motor}& \textbf{pers.}&\textbf{pott}&\textbf{sheep}&\textbf{sofa}&\textbf{train}&\textbf{tv.}\\
\tabucline[0.8pt]{-} 

PSANet&ResNet-101& 79.25&94.8&91.2&43.7&89.8&76.0& 80.6&94.4&88.4&93.3&44.5&88.6&56.0&89.4&87.5&85.8&88.2&65.0&92.2&50.7&86.7&77.4\\
+EPL&ResNet-101&\textbf{80.33}&95.1&92.0&44.4&90.5&\textbf{77.9}&\textbf{82.1}&94.5& \textbf{90.1}&\textbf{94.4}&43.8&89.0&\textbf{59.2}&90.1&88.1&\textbf{88.0}&88.9&65.0&91.9&\textbf{53.9}&\textbf{89.3}&\textbf{78.9}\\
\hline
PSPNet&ResNet-101& 79.50&94.9&90.8&44.2&90.0&74.0& 81.0&95.3&90.2&94.2&42.8&87.9&57.0&89.5&87.2&89.8&88.3&65.7&91.3&47.3&88.7&79.3\\
+EPL&ResNet-101&\textbf{80.46}&\textbf{95.1}&92.4&44.7&88.9&\textbf{75.7}&81.7&95.8& \textbf{91.5}&94.6&43.5&\textbf{89.1}&\textbf{59.4}&\textbf{90.5}&\textbf{88.3}&90.4&88.4&64.7&\textbf{93.4}&\textbf{49.0}&\textbf{91.5}&76.9\\
\hline 
DeepLab&ResNet-101&79.15&93.9&89.7&42.3&90.4&69.0& 81.1&93.0&91.0&93.5&41.8&90.2&62.1&90.8&88.6&86.4&86.8&67.0&87.5&50.3&87.5&79.2\\
+EPL&ResNet-101&\textbf{80.77}&94.2&89.8&\textbf{43.5}&90.6&\textbf{72.9}&81.8&93.4& 90.3&93.9&\textbf{47.2}&\textbf{92.3}&\textbf{64.7}&\textbf{91.8}&89.2&\textbf{88.7}&87.7&\textbf{70.4}&\textbf{89.9}&\textbf{59.8}&87.8&77.3\\
\hline
\end{tabu}
\end{center}
\caption{Category-level mIoU comparison on Pascal Voc 2012 Validation set.}
\label{tb:perclass_pascal_PSPNET}
\end{table*}
\subsection{Comparison with related works}
\label{section:c_w_r}
The section compares $L_{point}$ and $L_{line}$ with their related works. Note that none of the compared losses has been adapted to the multi-class segmentation problem in previous studies.

\textbf{Point loss vs. boundary loss:} A principal property of the potential fields is that points' potential energy increase with their spatial distance to the semantic boundary, conforming to the character of the distance fields. We consider the boundary loss \cite{bound_loss} related to our method and then assemble it to PSPNet (see Appendix \ref{section: append_comparison}). To make a fair comparison, we test the boundary loss $L_{bd}$ and $L_{point}$ with five loss weights $\{0.05,0.10,0.20,0.25, 0.50\}$, and then report both losses' best segmentation performances on two datasets. In Table \refTab{tb:comparison_related}, we see $L_{point}$ outperforms $L_{bd}$ because the boundary loss is not applicable for the multi-class task and is less informative than $L_{point}$ that conducts field regression in all directions (see Table \refTab{tab:comparison_bd} in Appendix).

\textbf{Equipotential line loss vs. dice loss:} This part compares the $L_{line}$ with the dice loss ($L_{dice}$) that has a similar optimization principle. We see $L_{dice}$ as an auxiliary for image segmentation and adopt the same evaluation protocol as experiments with $L_{point}$ (see Appendix \ref{section: append_comparison} and Table \refTab{tab:comparison_dice} for implementation details and more comparisons). In the end, we report the comparison results in Table \refTab{tb:comparison_related} and observe that $L_{line}$ performs better than $L_{dice}$ \cite{diceloss}. This comparison result proves the effect of anisotropic convolution and indicates that contour learning can benefit semantic segmentation.

\begin{table}
\centering
\small
\begin{tabu}[t]{m{1.5cm}<\centering|m{4cm}<\centering|m{1.2cm}<\centering}

\textbf{Dataset} & \textbf{Method} & \textbf{mIoU(\%)} \\

                                        \tabucline[0.8pt]{-}  
\multirow{5}{*}{Pascal Voc} & $+L_{ce}$ & 73.52 \\

~&$+L_{ce}+L_{bd}$ & 74.55\\

~&\textbf{$+L_{ce}+L_{point}$} &\textbf{74.84\tiny{(+1.32)}}\\

~& $+L_{ce}+L_{dice}$ & 74.45\\

~&\textbf{ $+L_{ce}+L_{line}$} & \textbf{74.71\tiny{(+1.19)}}\\
\hline
\multirow{5}{*}{Cityscapes} & $+L_{ce}$& 71.66 \\

~&$+L_{ce}+L_{bd}$& 71.81\\

~&\textbf{ $+L_{ce}+L_{point}$}&\textbf{72.40\tiny{(+0.74)}}\\

~&$+L_{ce}+L_{dice}$ & 70.30\\
~&\textbf{$+L_{ce}+L_{line}$} &\textbf{73.26\tiny{(+1.60)}}\\
\hline

\end{tabu}
\caption{Comparisons with the boundary loss \cite{bound_loss} ($L_{bd}$) and dice loss \cite{diceloss} (\textbf{$L_{dice}$ }) on two datasets' validation set.}
\label{tb:comparison_related}
\end{table}

\subsection{Main Results}
\label{main_results}
After loss balance, we apply the full EPL on five FCN baselines: PSPNet\cite{pspnet}, PSANet\cite{psanet}, DeepLab V3+\cite{Deeplabv3plus}, CCNet\cite{ccnet} and GSCNN\cite{gscnn}. Besides, we conduct intensive experiments with variable backbones: ResNet-50/101 \cite{ResNet}, MobileNet \cite{mobilenets}, DRN \cite{DRN} and WRNet38 \cite{WIDERESNET}. Note that all models are trained in a batch of  16  with  reported  image  size  in  Table \refTab{tb:main_pascal} \& \refTab{tb:main_city}.

\begin{figure}[h!]
\begin{center}
\includegraphics[width=8.5cm]{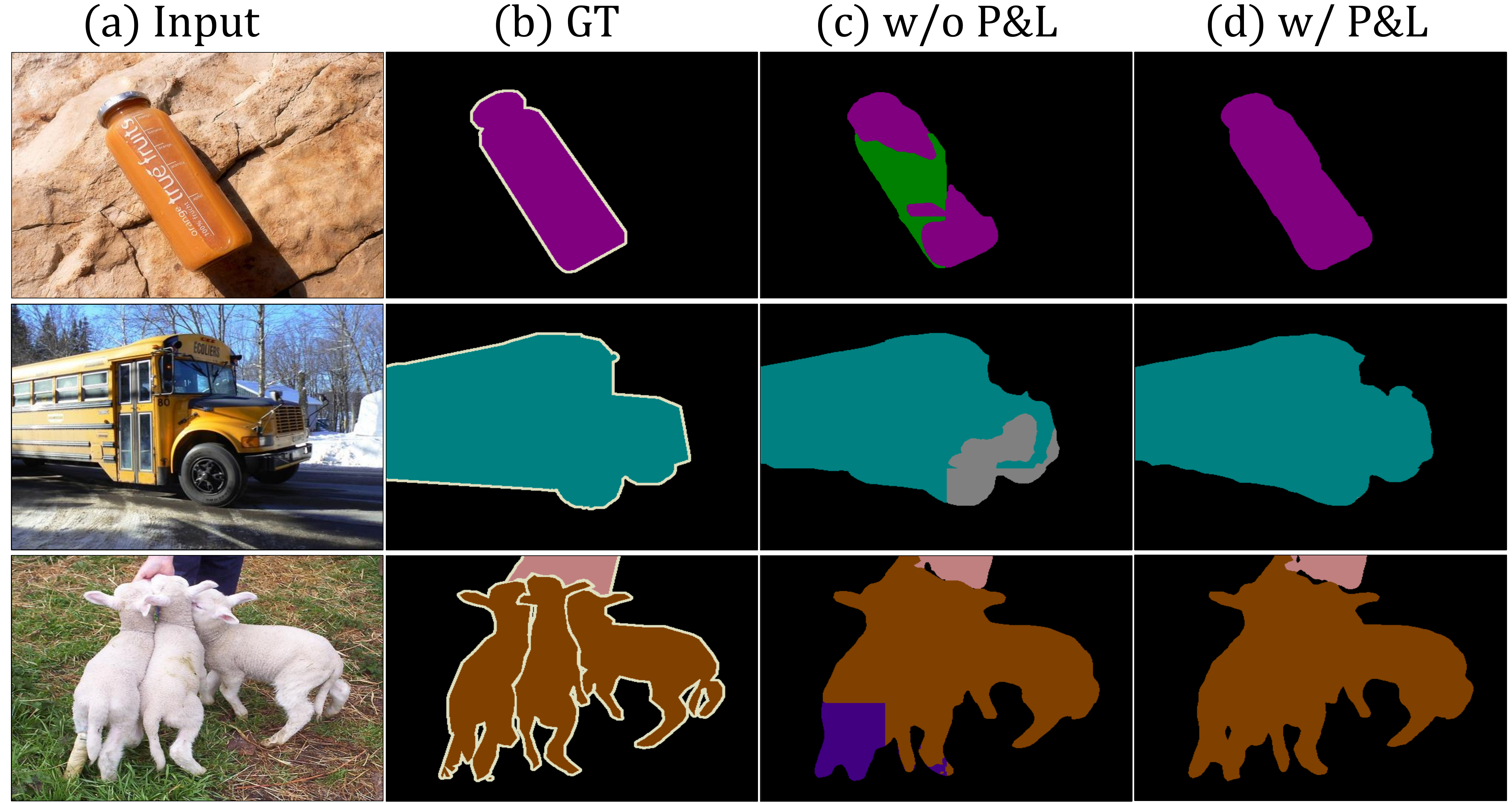}
\end{center}
   \caption{Qualitative comparison for segmentation results on Pascal Voc 2012 validation set.
   }
   \label{fig:Result_comparison_voc}
        
\end{figure}

\textbf{Performance on Pascal Voc 2012:} In Table \refTab{tb:main_pascal}, we compare baseline models' performances with and without using EPL. Models with EPL consistently outperform their corresponding baseline models with considerable mIoU increasements. In the best case, we improve the mIoU of DeepLab v3+ (with ResNet101) up to 1.62\%. \refFig{fig:Result_comparison_voc} visualizes the segmentation results, and we see that EPL can greatly help existing FCN-based segmentation models solve the challenges from the inter-class (the first and the second row) or the intra-class regions (the third row).

\begin{figure}[t]
\begin{center}
\includegraphics[width=8.5cm]{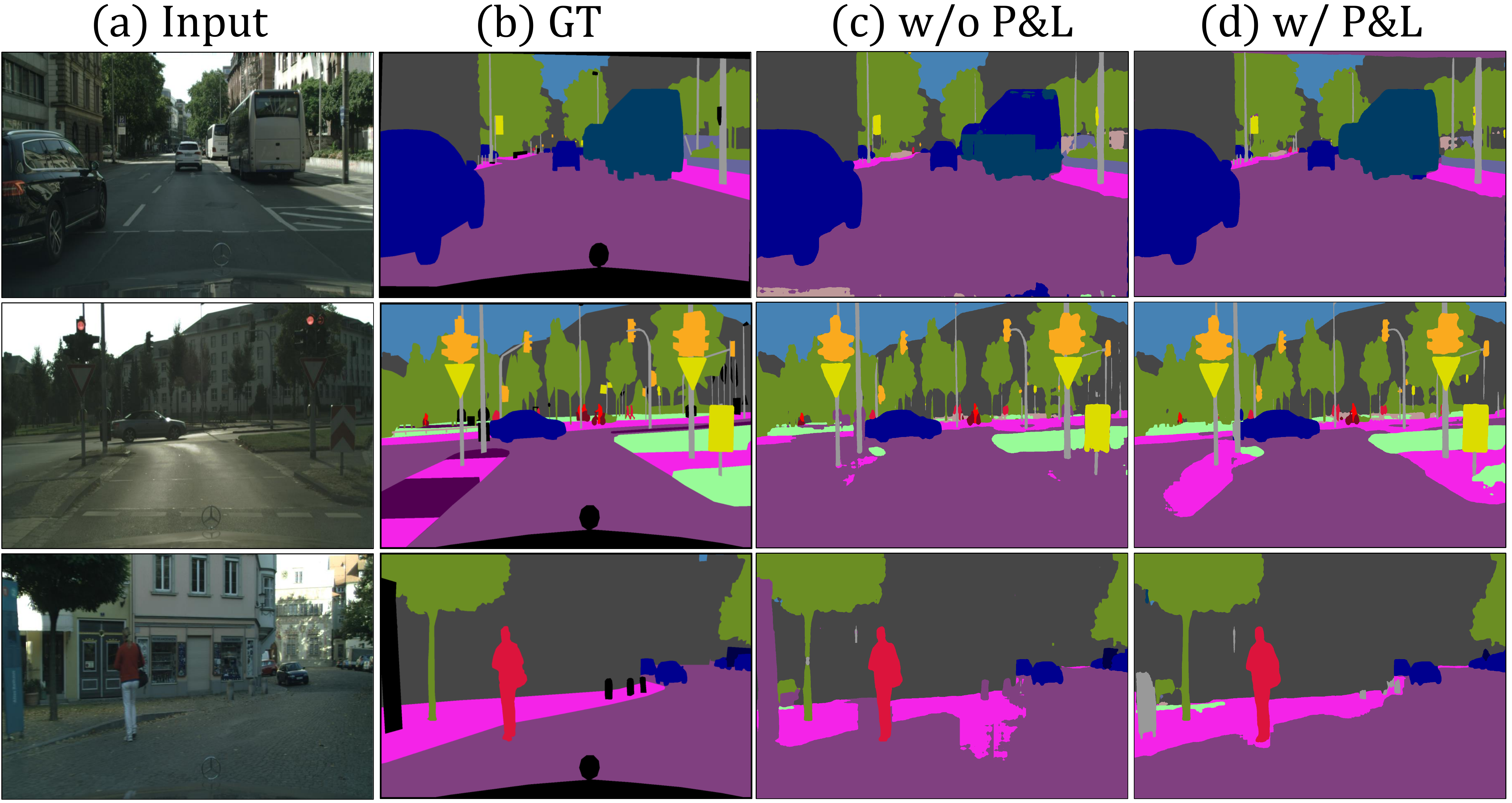}
\end{center}
   \caption{ Qualitative comparison for segmentation results on Cityscapes validation set.
  }
   \label{fig:Result_comparison_city}
  
\end{figure}

\textbf{Performance on Cityscapes:} We report the result on cityscapes in Table \refTab{tb:main_city}. Once again, we achieve substantial performance gains over all three FCN baseline models when we employ EPL, regardless of the backbone type. Similarly, visual comparisons are exhibited in \refFig{fig:Result_comparison_city}.

\textbf{Category-level evaluation.} We show the category-level mIoU comparison on both datasets before and after adopting EPL in Table \ref{tb:perclassL_city_PSPNET} \& \ref{tb:perclass_pascal_PSPNET}. On Cityscapes, we observe that EPL significantly improves baselines' ability to distinguish category pairs that have similar semantics or strong semantic relationships, such as \{``person", ``rider,"\}, \{``rider", ``bicycle"\}, and \{``traffic sign," ``traffic light"\}. On Pascal Voc 2012, EPL enhances baseline models' ability to segment categories with complicated shapes, such as the cow, dog, and dining table.
\begin{table}[h!]
\small
\begin{center}
\begin{tabu}{m{1cm}<{\centering}|m{1.2cm}<{\centering}|m{1.5cm}<{\centering}|m{1.2cm}<{\centering}|m{1.2cm}<{\centering}}

\textbf{Model}&\textbf{Size}&\textbf{Backbone}&\textbf{Method}&\textbf{mIoU(\%)}\\
\tabucline[0.8pt]{-} 

 \multirow{4}{*}{PSANet}&\multirow{4}{*}{465$\times$465}&\multirow{2}{*}{ResNet-50}&Baseline&77.85\\

 ~&~&~&+EPL&\textbf{79.08\tiny{(+1.23)}}\\

 ~&~&\multirow{2}{*}{ResNet-101}&Baseline&79.25\\

  ~&~&~&+EPL&\textbf{80.33\tiny{(+1.08)}}\\
\hline

 \multirow{4}{*}{PSPNet}&\multirow{4}{*}{473$\times$473}&\multirow{2}{*}{ResNet-50}&Baseline&78.02\\

 ~&~&~&+EPL&\textbf{78.84\tiny{(+0.82)}}\\

 ~&~&\multirow{2}{*}{ResNet-101}&Baseline&79.50\\

  ~&~&~&+EPL&\textbf{80.46\tiny{(+0.96)}}\\
\hline

 \multirow{6}{*}{DeepLab}&\multirow{6}{*}{513$\times$513}&\multirow{2}{*}{MoblieNet}&Baseline&71.49\\

 ~&~&~&red{+EPL}&\textbf{72.65\tiny{(+1.16)}}\\

 ~&~&\multirow{2}{*}{DRN-54}&Baseline&79.58\\

  ~&~&~&+EPL&\textbf{80.63\tiny{(+1.05)}}\\

 ~&~&\multirow{2}{*}{ResNet-101}&Baseline&79.15\\

  ~&~&~&+EPL&\textbf{80.77\tiny{(+1.62)}}\\
\hline

\end{tabu}
\end{center}

\caption{Overall results on Pascal Voc validation set.}
\label{tb:main_pascal}
\end{table}

\begin{table}[h!]
\small
\begin{center}
\begin{tabu}{m{1cm}<{\centering}|m{1.2cm}<{\centering}|m{1.5cm}<{\centering}|m{1.2cm}<{\centering}|m{1.2cm}<{\centering}}
\textbf{Model}&\textbf{Size}&\textbf{Backbone}&\textbf{Method}&\textbf{mIoU(\%)}\\
\tabucline[0.8pt]{-} 

 \multirow{4}{*}{PSANet}&\multirow{4}{*}{560$\times$560}&\multirow{2}{*}{ResNet-50}&Baseline&76.69\\
 ~&~&~&+EPL&\textbf{77.61\tiny{(+0.92)}}\\

 ~&~&\multirow{2}{*}{ResNet-101}&Baseline&78.01\\

  ~&~&~&+EPL&\textbf{79.46\tiny{(+1.45)}}\\
\hline

 \multirow{4}{*}{PSPNet}&\multirow{4}{*}{560$\times$560}&\multirow{2}{*}{ResNet-50}&Baseline&77.34\\

 ~&~&~&+EPL&\textbf{78.49\tiny{(+1.15)}}\\

 ~&~&\multirow{2}{*}{ResNet-101}&Baseline&78.35\\

  ~&~&~&+EPL&\textbf{79.44\tiny{(+1.09)}}\\
\hline

 \multirow{2}{*}{DeepLab}&\multirow{2}{*}{560$\times$560}&\multirow{2}{*}{WRNet-38}&Baseline&79.38\\
  ~&~&~&+EPL&\textbf{80.34\tiny{(+0.96)}}\\
  \hline
   \multirow{2}{*}{CCNet}&\multirow{2}{*}{560$\times$560}&\multirow{2}{*}{WRNet-38}&Baseline&77.73\\
  ~&~&~&+EPL&\textbf{78.81\tiny{(+1.08)}}\\
  \hline
   \multirow{2}{*}{GSCNN}&\multirow{2}{*}{560$\times$560}&\multirow{2}{*}{WRNet-38}&Baseline&80.67\\
   
  ~&~&~&+EPL&\textbf{81.78\tiny{(+1.11)}}\\
\hline
\end{tabu}
\end{center}
\caption{Overall results on Cityscapes validation set.}
\label{tb:main_city}
\end{table}

\subsection{Comparison with state-of-the-art methods}
In this section, we compare the proposed EPL module with the existing boundary segmentation approaches \cite{pointrend,segfix,DSN} in Cityscapes. In Table \refTab{tb:main_pascal} and \refTab{tb:main_city}, we observe that EPL enhances all segmentation baselines by over 1\% regardless of the backbone models. Compared (in Table \refTab{tb:comparison_sota}) with other boundary segmentation studies \cite{pointrend,DSN,segfix,inf}, EPL is slightly worse than InverseForm (\textbf{InF}) yet competitive against the other studies. Besides, we compare the properties of each method in Table \refTab{tb:property_comparison} and observe that EPL is the only approach that does not utilize edge maps for the network training and brings no increased parameter size and inference cost.
\begin{table}[h!]
\small
\begin{center}
\begin{tabu}{m{3.5cm}|m{1.6cm}<{\centering}|m{1.3cm}<{\centering}}
\textbf{Method}&\textbf{Backbone}&\textbf{mIoU(\%)}\\
\tabucline[0.8pt]{-} 
PSPNet&ResNet-50&77.34\\
PSPNet w/ EPL&ResNet-50&\textbf{78.49\tiny{(+1.15)}}\\
\hline 
PSANet&ResNet-101&78.01\\
PSANet w/ EPL&ResNet-101&\textbf{79.46\tiny{(+1.45)}}\\
\hline 
DeepLabv3 &ResNet-101&77.80\\
DeepLabv3 w/ PR\cite{pointrend} & ResNet-101&78.40\tiny{(+1.20)}\\
\hline 

DeepLabv3 &ResNet-50&79.18\\
DeepLabv3 w/ IABL\cite{IABL}&ResNet-50&79.94\tiny{(+0.76)}\\
\hline
DeepLabv3+&WRNet-38&79.50\\
DeepLabv3+ w/ SFix\cite{segfix}&WRNet-38&80.30\tiny{(+0.80)}\\
\hline
DeepLabv3+&ASPP\cite{Deeplabv3plus}&81.30\\
DeepLabv3+ w/ DSN\cite{DSN}&ASPP\cite{Deeplabv3plus}&82.40\tiny{(+1.10)}\\
\hline
DeepLabv3+ (our Impl.)&WRNet-38&79.38\\
DeepLabv3+ w/ EPL&WRNet-38&\textbf{80.34\tiny{(+0.96)}}\\
\hline
GSCNN &WRNet-38&81.0\\GSCNN w/ InF \cite{inf} & WRNet-38&82.60\tiny{(+1.50)}\\
\hline
GSCNN (our Impl.)&WRNet-38&80.67\\
GSCNN w/ EPL&WRNet-38&\textbf{81.78\tiny{(+1.11)}}\\
\hline

\end{tabu}
\end{center}
\caption{Comparisons with SOTA studies on Cityscapes. Note that our result (\textbf{our Impl.}) is experimented with  $560\times 560$ images and hence results in a small mIoU gap. }
\label{tb:comparison_sota}

\end{table}

\begin{table}

\centering
\small

\begin{tabu}[t]{m{2.0cm}|m{1.7cm}<\centering|m{1.7cm}<\centering|m{1.7cm}<\centering}

\textbf{Method} & \textbf{No
param. overhead} & \textbf{No edge super.}& \textbf{No infer. overhead}\\

                                        \tabucline[0.8pt]{-}
                                        SFix \cite{segfix} &{-}&{-}&{-}\\
                                        
DSN \cite{DSN}&{-}&{-}&{-}\\ 
PR\cite{pointrend} &{-}&{\checkmark}&{-}\\
InF\cite{inf}&{-} &{-}&{\checkmark}\\
IABL\cite{IABL}&{-}&{\checkmark}&{\checkmark}\\
\hline
EPL&{\checkmark}&{\checkmark}&{\checkmark}\\

\hline

\end{tabu}

\caption{Comparisons of the SOTA boundary segmentation methods. We evaluate these studies' properties from three perspectives: whether using edge maps for boundary supervision (\textbf{super.}) and increasing parameters (\textbf{param.}) and inference (\textbf{infer.}) overhead.}
\label{tb:property_comparison}
     
\end{table}

\section{Conclusion}
\label{sec:conclusion}
This paper addresses the semantic boundary segmentation problem with anisotropic field regression and category-level contour learning. With the proposed EPL (equipotential learning) module, we transfer the original probability estimation problem to the self-defined potential domain with the anisotropic convolution. Besides, we sequentially introduced the point loss to fit the image content along different directions from variable distances and the equipotential line loss to enforce the category-level contour learning. Experiments on Pascal Voc 2012, Cityscapes show that the designed EPL, serving as an add-on method, can significantly enhance existing FCN models' performance on the semantic boundary regions. Compared with other studies \cite{pointrend,efficient,segfix,DSN,inf}, our approach does not introduce additional supervision information and adds no parameters and inference cost. We believe that EPL can be generalized and benefit other segmentation tasks, like point cloud \cite{pcsg1,pcsg2}, instance, and panoptic segmentation \cite{ps1}.

{\appendix
In this part, we report more comparison results of our approach in ablation experiments. In the end, we present more quantitative results to support our paper.
\section{Detailed ablation results}
This section presents more experiment details and comparison results for the ablation experiment in Section \ref{section:abl_exp}.

\subsection{Comparisons results of different splitters:} We experiment AC with with three different splitters, $A, B$ and $C$ (shown in \refFig{fig:example_of_ac}), that contain different directional vectors. Next, we compare the segmentation performance of the point loss and the equipotential line loss on PSPNet when using different splitters. As reported detailed comparison results in Table \refTab{tab:appendix_point_abl} \& \refTab{tab:appendix_line_abl}, we see that $L_{point}$ and $L_{line}$ perform best when using the splitter $A$.

\begin{table}[h]
\small
\begin{tabu}{m{1.5cm}<{\centering}|m{1.0cm}<{\centering}|m{0.8cm}<\centering|m{0.8cm}<\centering|m{0.8cm}<\centering|m{0.8cm}<\centering}

\textbf{Dataset} &\textbf{Method} &\textbf{Kernel} &\multicolumn{3}{c}{\textbf{mIoU (\%)}}\\
\tabucline[0.8pt]{-} 
\multirow{8}{*}{Pascal Voc}&\multirow{2}{*}{Baseline}&\multirow{2}{*}{-}&\multicolumn{3}{c}{73.52}\\
~&~&~&A&B&C\\
~&\multirow{3}{*}{+Point$^1$}
&7&74.07  &	74.24 &73.80 \\
~&~&9&73.85 &74.36 &	73.83 \\
~&~&11&\textbf{74.08} &\textbf{74.55}&\textbf{74.35} \\
\cline{2-6}
~&\multirow{3}{*}{+Point$^2$}
&7&74.09 &\textbf{74.42} &\textbf{74.63} \\
~&~&9&\textbf{74.42} &73.60 &71.21 \\
~&~&11&74.21 &72.96 &74.29 \\

\hline
\multirow{8}{*}{Cityscapes}&\multirow{2}{*}{Baseline}&\multirow{2}{*}{-}&\multicolumn{3}{c}{71.66}\\
~&~&~&A&B&C\\
~&\multirow{3}{*}{+Point$^1$}
&9&	70.44 &	\textbf{72.07} &	61.37 \\
~&~&11&	70.96 &	55.04 &	67.01 \\
~&~&13&	\textbf{72.71} &	66.61 &	\textbf{72.22} \\
\cline{2-6}
~&\multirow{3}{*}{+Point$^2$}
&9&\textbf{72.39} &	71.50 &	68.25 \\
~&~&11&70.93 &	\textbf{73.25} &	\textbf{73.07} \\
~&~&13&	71.02 &	64.72 &	71.78	\\

\hline
\end{tabu}
\caption{Detailed performance comparisons of the Point loss on Pascal Voc 2012 and Cityscapes validation set. The superscript of “+Point” denotes the norm type, and $A, B, C$ columns indicate that performances are obtained using the corresponding splitter.}
\label{tab:appendix_point_abl}
     
\end{table}
\begin{table}[h]
\small
\begin{tabu}{m{1.5cm}<\centering|m{1.0cm}<\centering|m{0.8cm}<\centering|m{0.8cm}<\centering|m{0.8cm}<\centering|m{0.8cm}<\centering}

\textbf{Dataset} &\textbf{Method} &\textbf{Kernel} &\multicolumn{3}{c}{\textbf{mIoU (\%)}}\\
\tabucline[0.8pt]{-} 
\multirow{5}{*}{Pascal Voc}&\multirow{2}{*}{Baseline}&\multirow{2}{*}{-}&\multicolumn{3}{c}{73.52}\\
~&~&~&A&B&C\\
~&\multirow{3}{*}{+Line}
&7&\textbf{74.15} &	73.50&	\textbf{73.94}\\
~&~&9&73.47&73.33&	73.73\\
~&~&11&74.05&\textbf{73.74}	&73.62\\

\hline
\multirow{5}{*}{Cityscapes}&\multirow{2}{*}{Baseline}&\multirow{2}{*}{-}&\multicolumn{3}{c}{71.66}\\
~&~&~&A&B&C\\
~&\multirow{3}{*}{+Line}
&9&	71.78&	\textbf{72.03}&	71.19\\
~&~&11&	\textbf{73.26}&	66.80&	\textbf{71.35}\\
~&~&13&	72.11&	60.42&	66.96\\

\hline
\end{tabu}
\caption{Detailed performance comparison of the equipotential line loss on Pascal Voc 2012 and Cityscapes validation set. $A, B, C$ denote the splitter type.}
\label{tab:appendix_line_abl}
\vspace{-5pt}
\end{table}

\subsection{Comparisons with related works}
\label{section: append_comparison}
In Section \ref{section:c_w_r}, we compared the point loss with the boundary loss,  the equipotential line loss, and the dice loss. Here, we report the detailed results of each comparison pair.
}

\textbf{Implementation details:} To apply the boundary loss to the segmentation network, we follow the practice of \cite{bound_loss}, firstly computing the ground-truth boundary-distance map of the true semantic labels and then integrating the boundary loss with the cross-entropy loss. Also, we adopt the same implementation protocol when comparing the dice loss with our equipotential line loss.

\textbf{Detailed comparison results:} For each comparison pair, we test 5 different loss weights and report the experiment details and conditions where the performances are achieved. We respectively set the splitter, kernel size, and loss norm of {‘A’, 7} and {‘A’, 11} when experimenting on Pascal Voc 2012 and cityscapes. Besides, the point loss is formulated as the $L_{2}$ norm consistently.

\begin{table}[t]
    \centering
    \small
    \begin{tabu}{cccc}
         \textbf{Dataset}&\textbf{weight}&\textbf{mIoU(B.\%)}&\textbf{mIoU(P.\%)}  \\
\tabucline[0.8pt]{-} 
\multirow{6}{*}{Pascal Voc}&Baseline&\multicolumn{2}{c}{73.52}\\

         ~&0.05&72.88&73.31\\
         ~&0.10&\textbf{74.55}&\textbf{74.84}\\
         ~&0.20&73.59&72.92\\
         ~&0.25&72.17&64.44\\
         ~&0.50&43.95&73.61\\
\hline
\multirow{6}{*}{Cityscapes}&Baseline&\multicolumn{2}{c}{71.66}\\

        ~&0.05&\textbf{71.81}&65.92\\
         ~&0.10&68.75&70.93\\
         ~&0.20&71.78&\textbf{72.40}\\
         ~&0.25&69.65&66.11\\
         ~&0.50&66.86&69.04\\
         \hline
    \end{tabu}
    \caption{Quantitative comparison between the boundary loss (\textbf{B.}) and our point loss (\textbf{P.}).}
    \label{tab:comparison_bd}
         
\end{table}

\begin{table}[t]
    \centering
    \small
    \begin{tabu}{cccc}

         \textbf{Dataset}&\textbf{weight}&\textbf{mIoU(D.\%)}&\textbf{mIoU(L.\%)}  \\
             \tabucline[0.8pt]{-} 
\multirow{6}{*}{Pascal Voc}&Baseline&\multicolumn{2}{c}{73.52}\\

         ~&0.05&72.88&73.49\\
         ~&0.10&74.07&73.68\\
         ~&0.20&\textbf{74.45}&\textbf{74.71}\\
         ~&0.25&74.36&68.22\\
         ~&0.50&73.45&72.83\\
         \hline
\multirow{6}{*}{Cityscapes}&Baseline&\multicolumn{2}{c}{71.66}\\
         ~&0.05&57.49&56.72\\
         ~&0.10&57.89&68.58\\
         ~&0.20&61.30&71.47\\
         ~&0.25&60.54&70.87\\
         ~&0.50&\textbf{70.30}&\textbf{73.26}\\
         \hline
    \end{tabu}
    \caption{Quantitative comparison between the dice loss (\textbf{D. }) and our equipotential line loss (\textbf{L.}) }
    \label{tab:comparison_dice}
     
\end{table}

\bibliography{main}
\bibliographystyle{IEEEtranS}
\end{document}


\title{Supplementary Material}

\markboth{Journal of \LaTeX\ Class Files,~Vol.~14, No.~8, August~2021}%
{Shell \MakeLowercase{\textit{et al.}}: A Sample Article Using IEEEtran.cls for IEEE Journals}


\maketitle
\section{Additional ablational results }
\begin{figure}[h]
    \centering
\includegraphics[width=8.5cm, height=4cm]{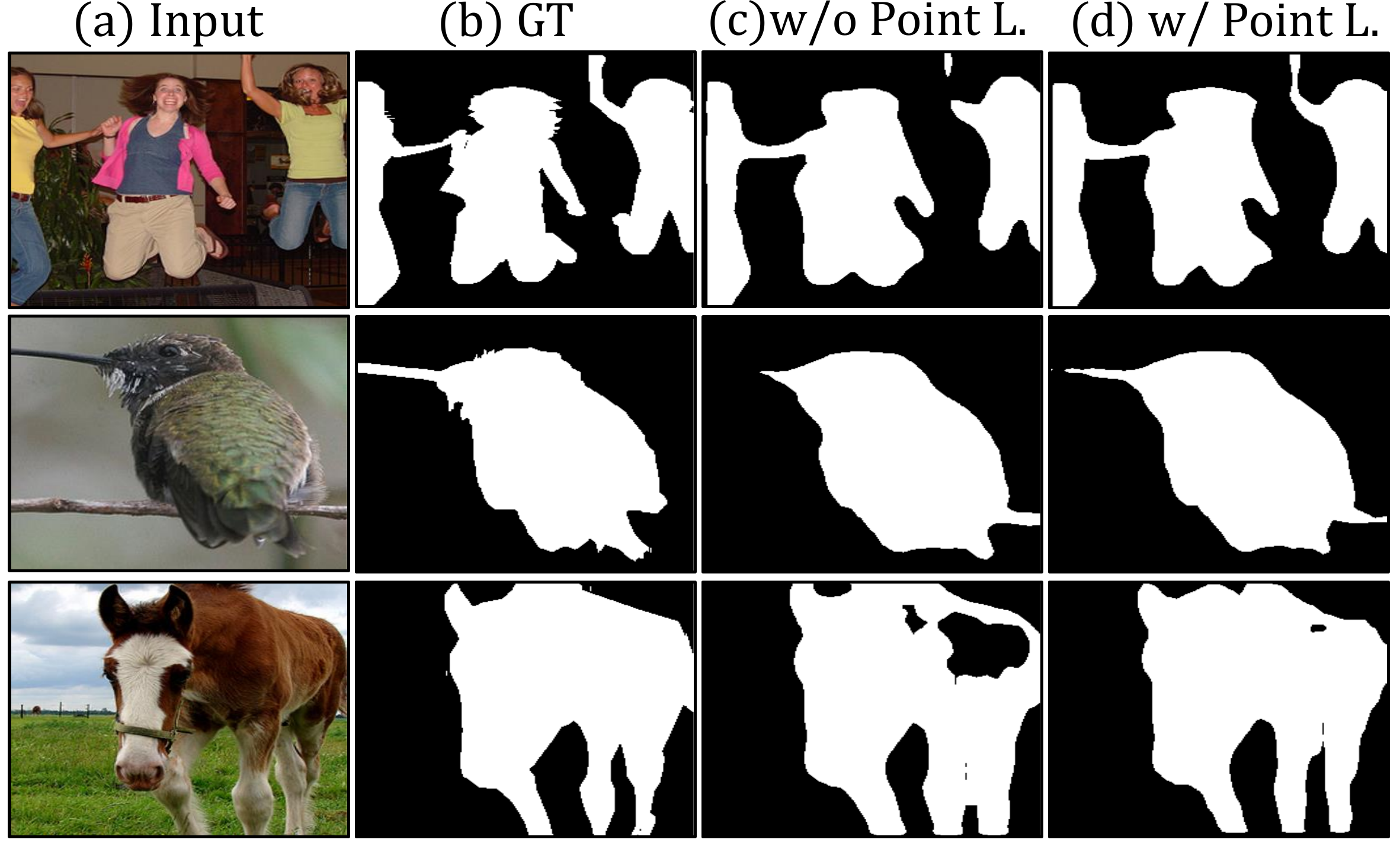}
    \caption{Example results of the point loss on Pascal Voc 2012. We respectively visualize the potential field of the ``person", ``bird", ``cow" category of the input image.}
    \label{fig:appendix_point_pascal}
    \vspace{-10pt}
\end{figure}

\begin{figure}[h]
    \centering
\includegraphics[width=8.5cm, height=4cm]{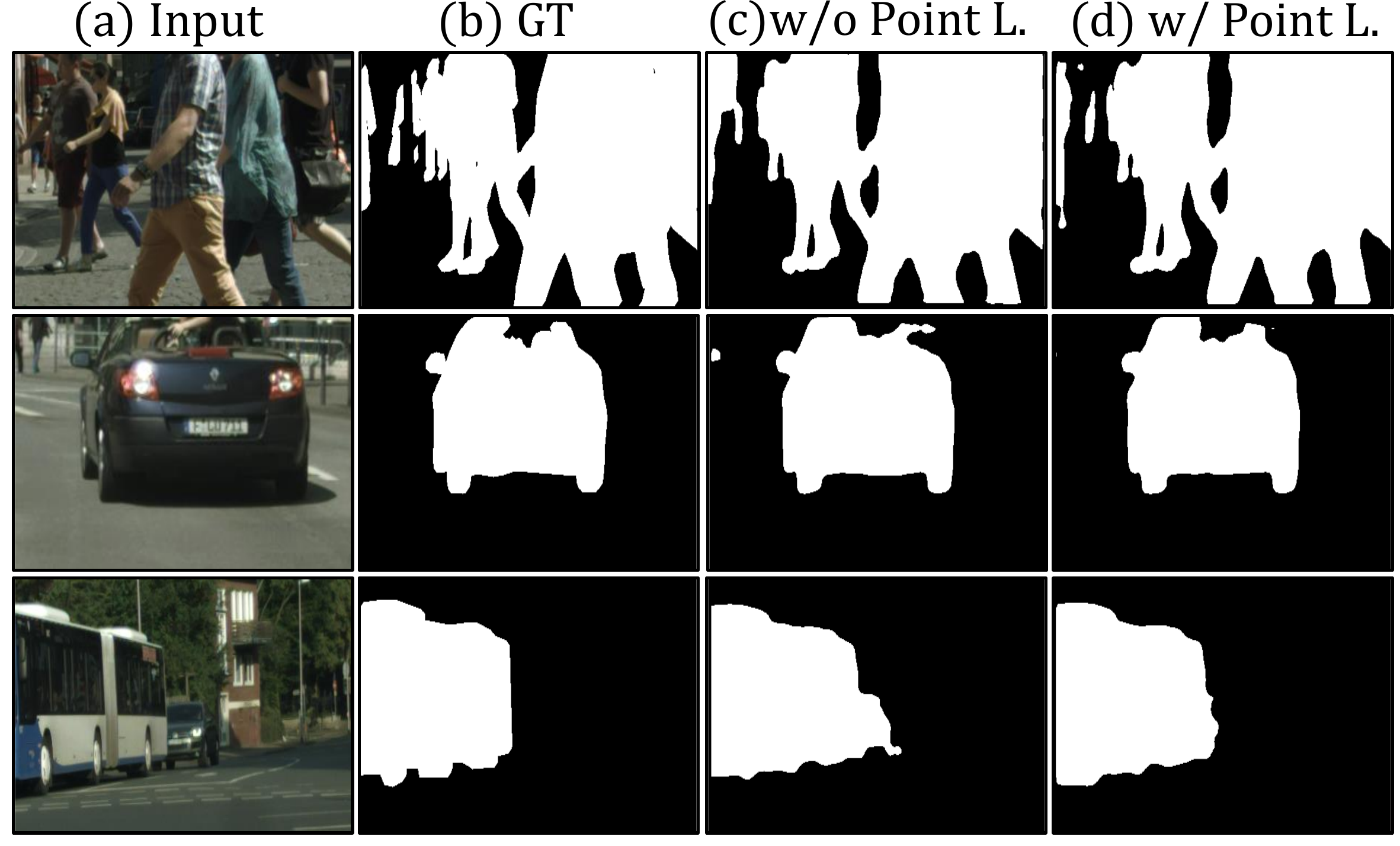}
    \caption{Example results of the point loss on Cityscapes. We respectively visualize the potential field of the ``human", ``car", ``bus" category of the input image.}
    \label{fig:appendix_point_city}
     \vspace{-10pt}
\end{figure}

\begin{figure}[h]
    \centering
\includegraphics[width=8.5cm, height=4cm]{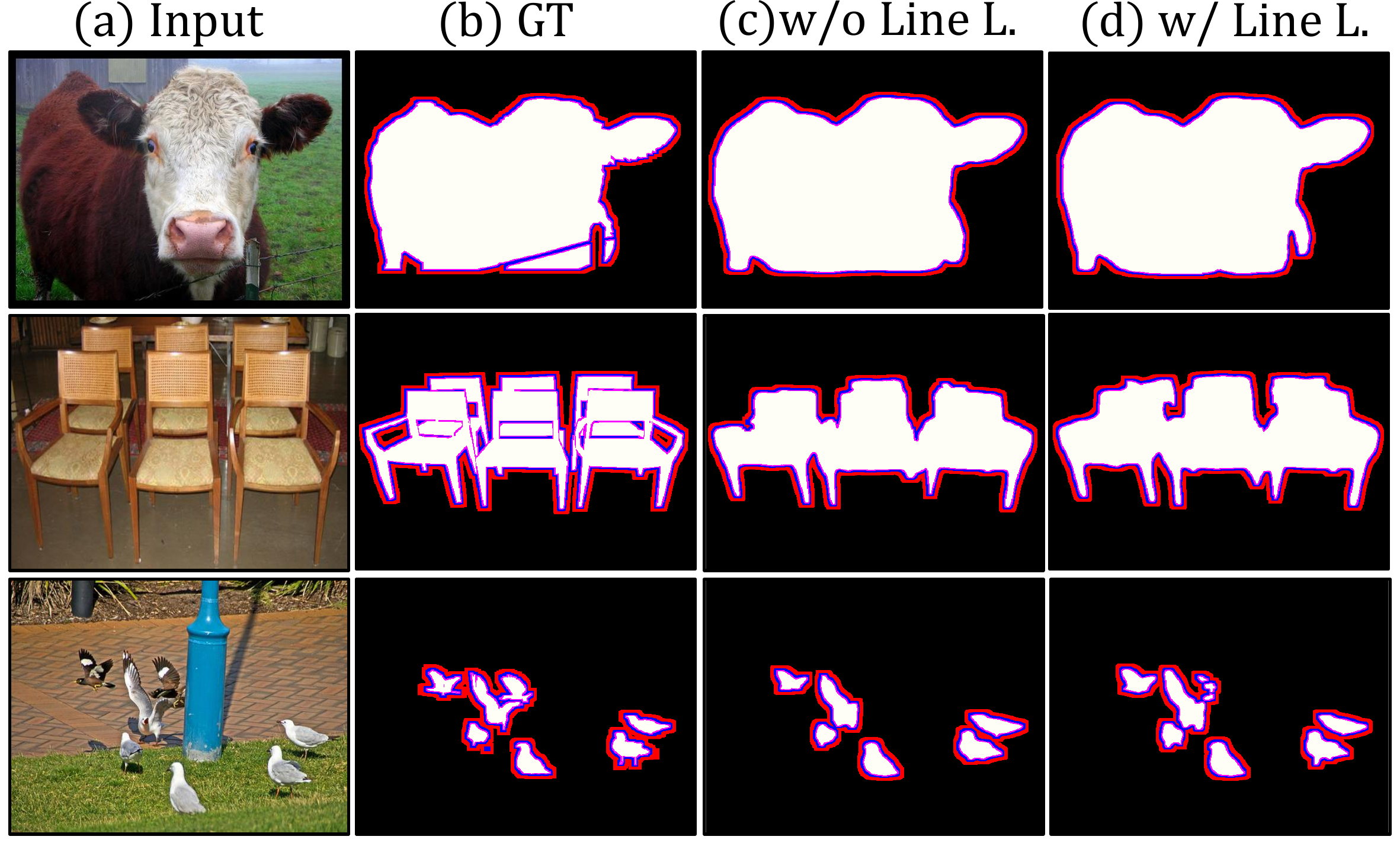}
    \caption{Example results of the equipotential line poss on Pascal Voc. We respectively visualize the equipotnetial line region of the ``cow", ``chair", ``bird" category of the input image.}
    \label{fig:appendix_line_pascal}
        \vspace{-10pt}
\end{figure}

\begin{figure}[h]
    \centering
\includegraphics[ width=8.5cm, height=4cm]{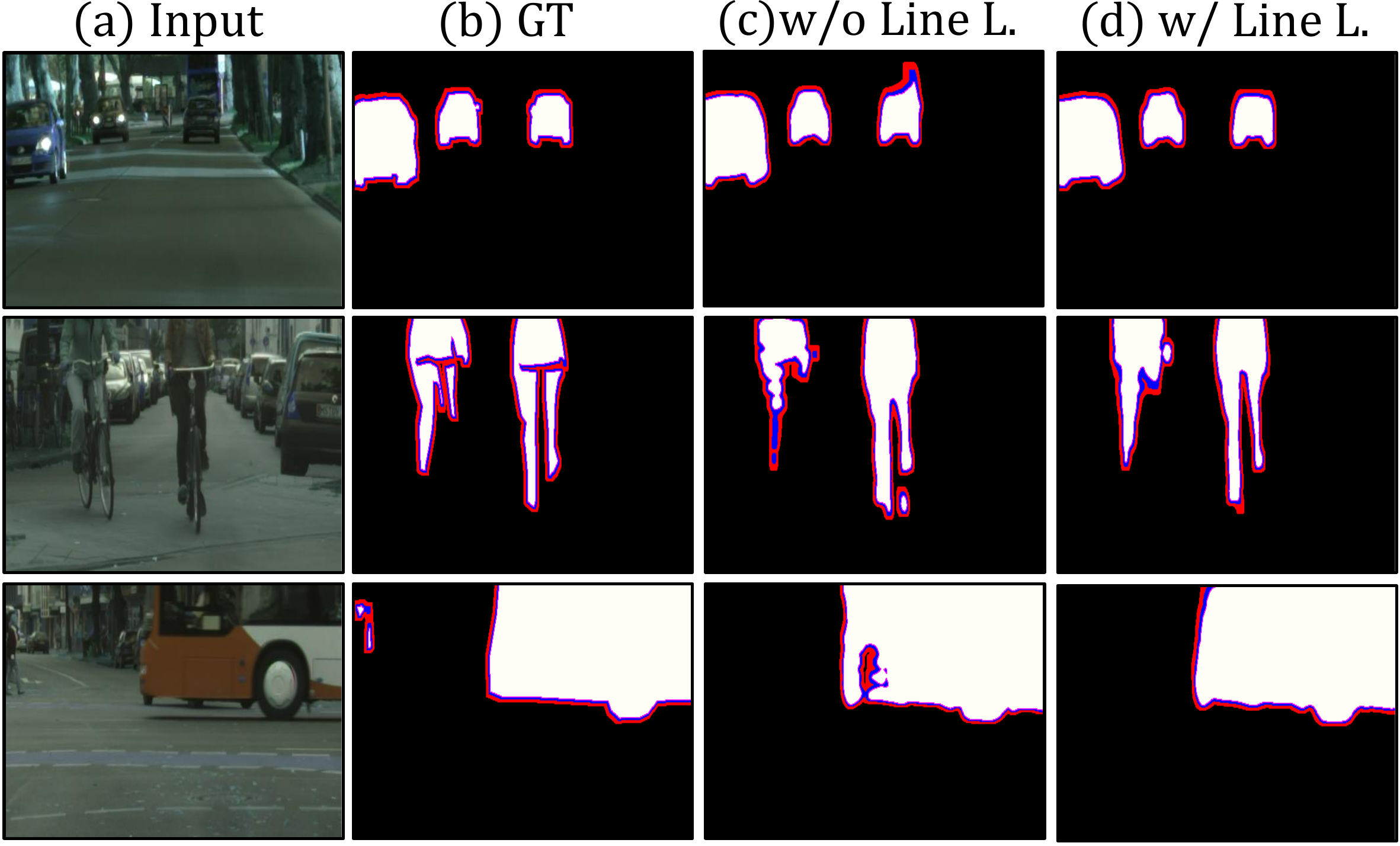}
    \caption{Example results of the equipotential line poss on Cityscapes.We respectively visualize the equipotential line region of the ``car", ``rider", ``bus" category of the inputs.}
    \label{fig:appendix_line_city}
    \vspace{-10pt}
\end{figure}

In this section, We present more comparison results of the ablation experiments, showing the impacts of $L_{point}$ and $L_{line}$ respectively. In  \refFig{fig:appendix_point_pascal} to \refFig{fig:appendix_point_city}, we visualize the category-level potential fields of the ground-truth, with/without our $L_{point}$ in 2-4th row. One can observe that our proposed point loss significantly improve the prediction consistency in the class channel and thus benefit the practical semantic segmentation. In \refFig{fig:appendix_line_pascal} to \refFig{fig:appendix_line_city}, we similarly visualize the category-level equipotential line region of the ground-truth, with/without our $L_{line}$, we see that the equipotential line loss $L_{line}$ enforce the category-level contour learning to help the network refine its prediction in the semantic boundary areas. 

\begin{figure}[h]
    \centering
    \includegraphics[width=8.5cm, height=4cm]{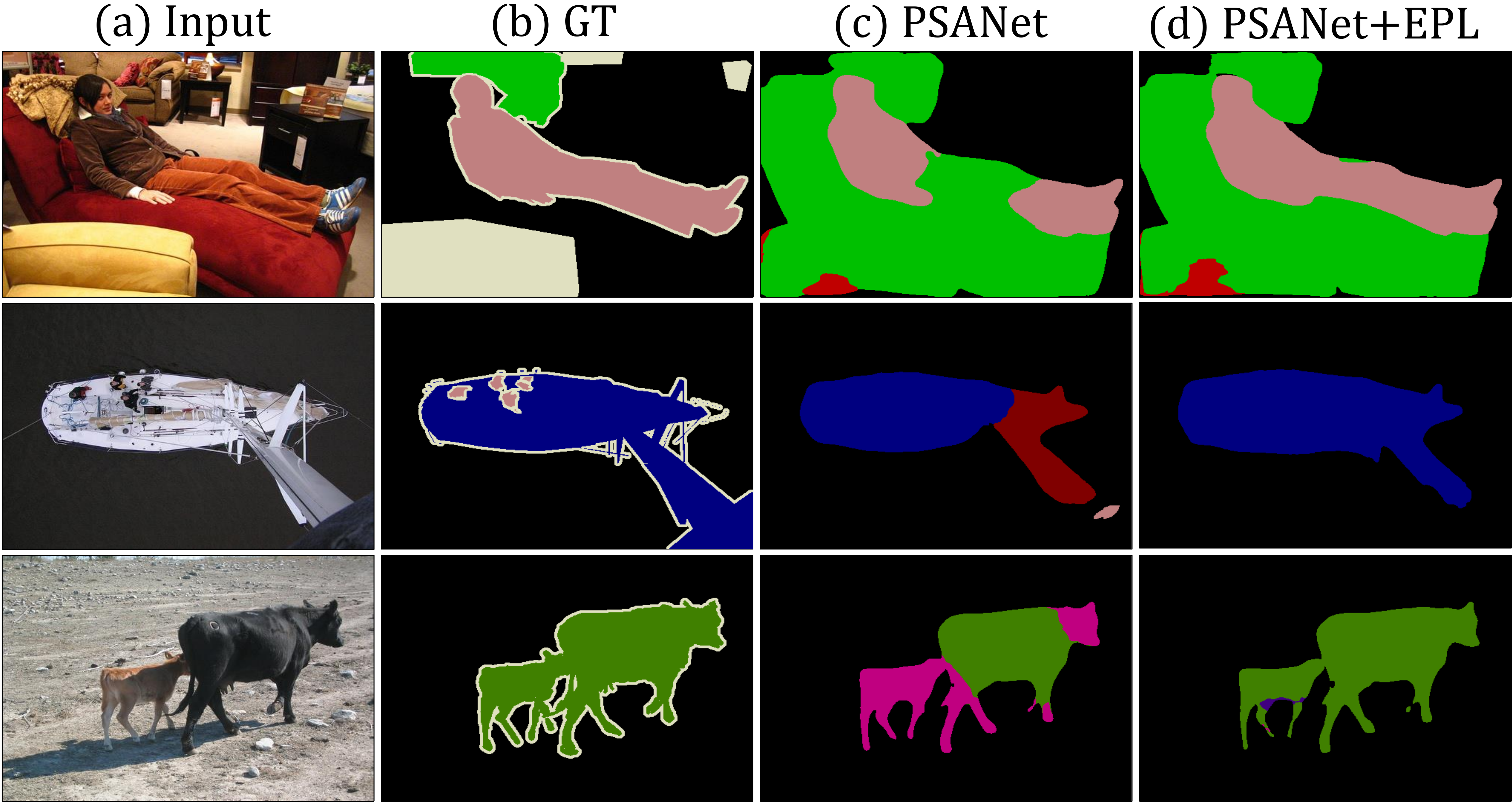}
    \caption{Qualitative comparison for segmentation results on Pascal VOC 2012 validation set. }
    \label{fig:appendix_pascal_psa}
    \vspace{-10pt}
\end{figure}
\begin{figure}[h]
    \centering
    \includegraphics[width=8.5cm, height=4cm]{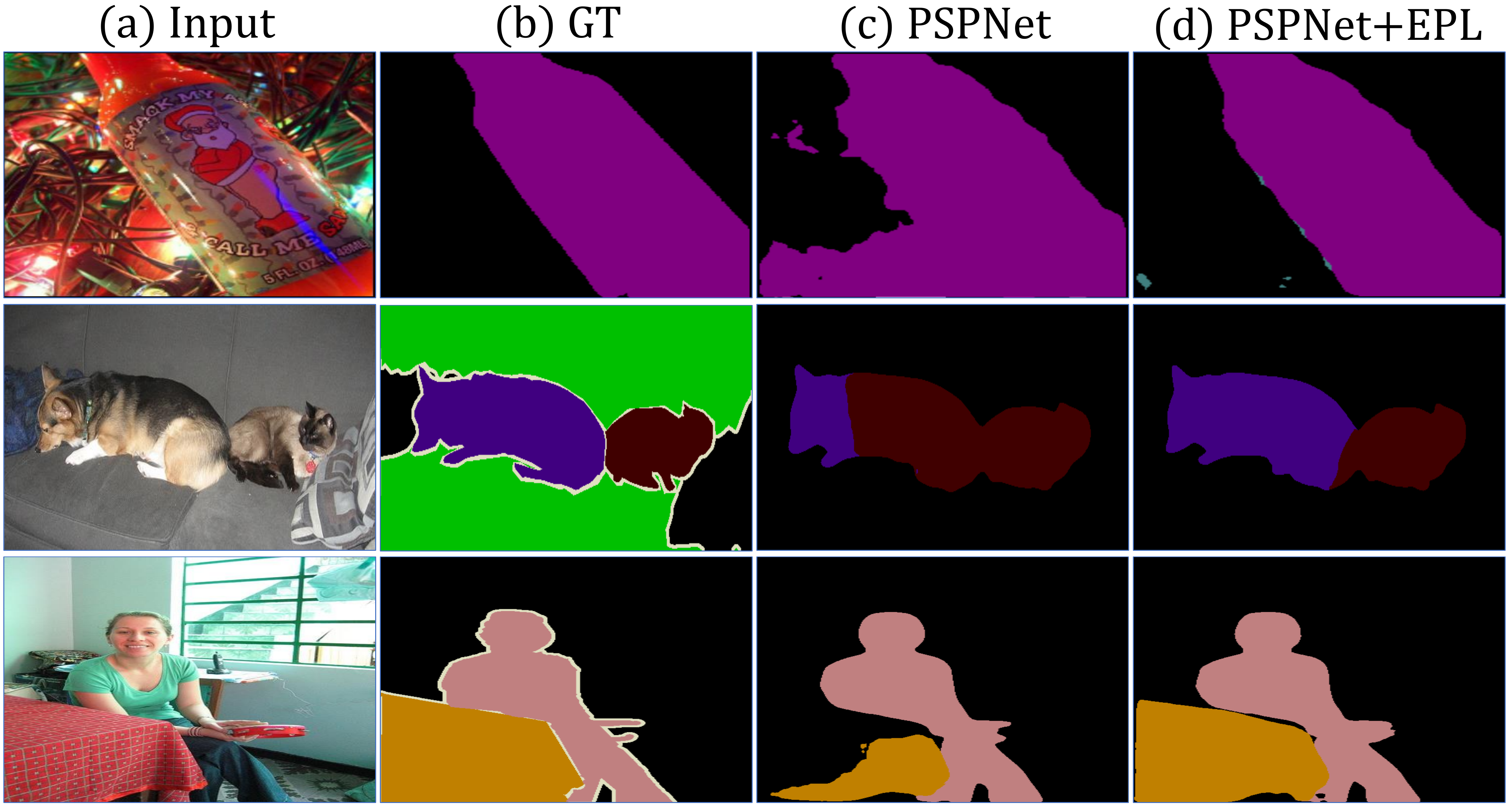}
    \caption{Qualitative comparison for segmentation results on Pascal VOC 2012 validation set.}
    \label{fig:appendix_pascal_psp}
    \vspace{-10pt}
\end{figure}
\begin{figure}[j!]
    \centering
    \includegraphics[width=8.5cm, height=4cm]{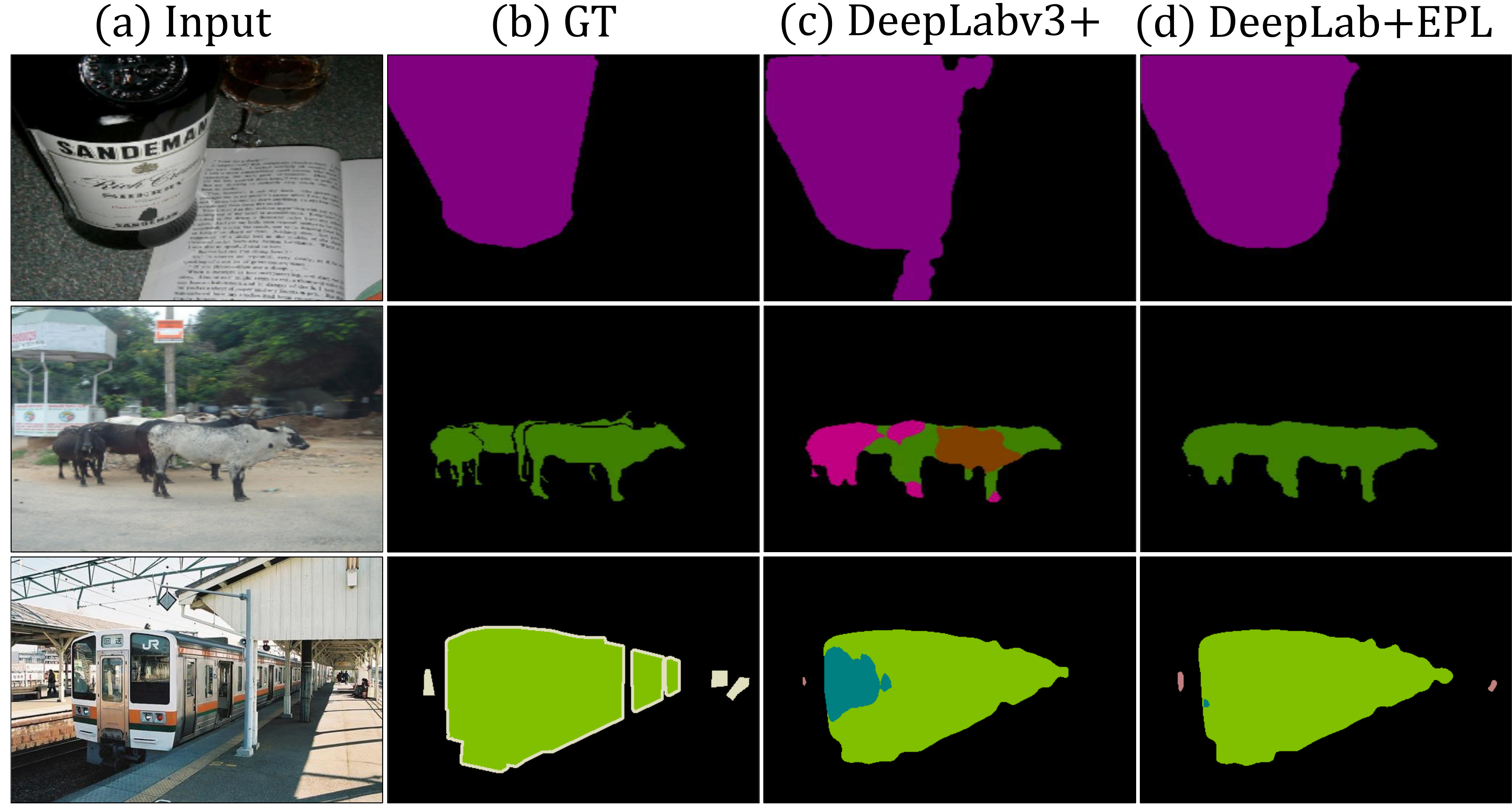}
    \caption{ Qualitative comparison for segmentation results on Pascal VOC 2012 validation set.}
    \label{fig:appendix_pascal_deep}
    \vspace{-10pt}
\end{figure}

\begin{figure}[h!]
    \centering
    \includegraphics[width=8.5cm, height=4cm]{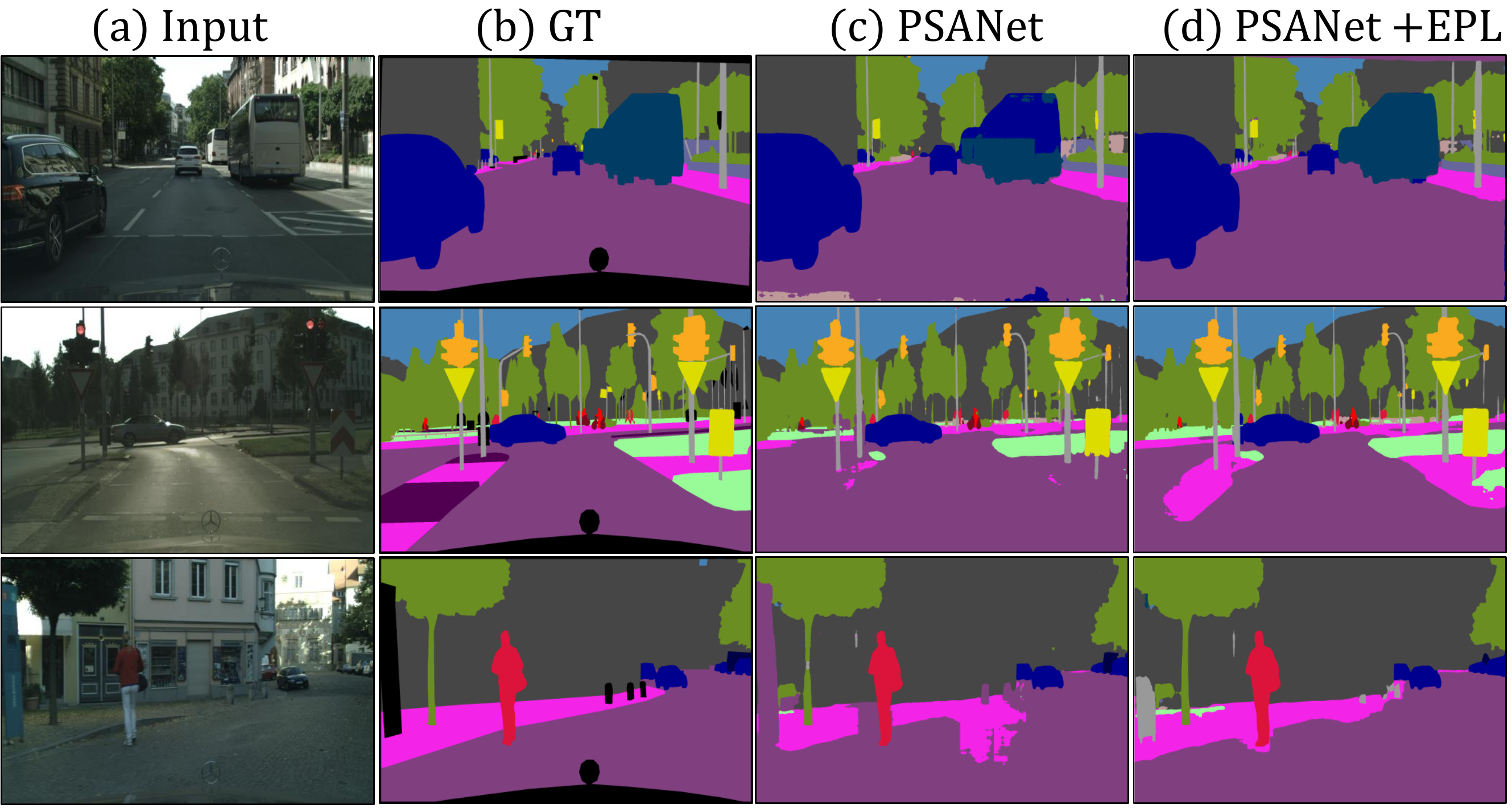}
    \caption{Qualitative comparison for segmentation results on Cityscapes validation set.}
    \label{fig:appendix_city_psa}
    \vspace{-10pt}
\end{figure}
\begin{figure}[h!]
    \centering
    \includegraphics[width=8.5cm, height=4cm]{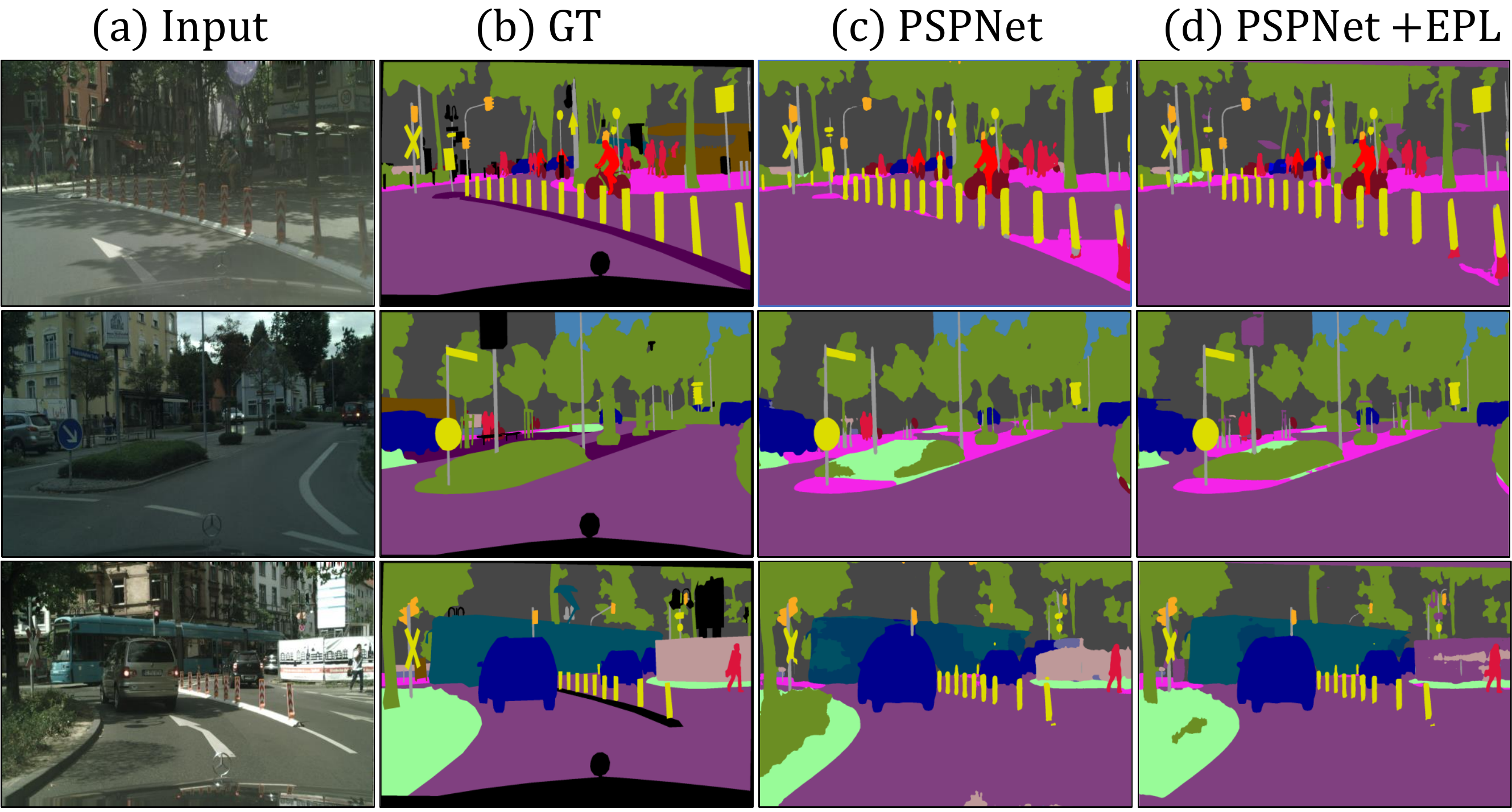}
    \caption{Qualitative comparison for segmentation results on Cityscapes validation set.}
    \label{fig:appendix_city_psp}
    \vspace{-10pt}
\end{figure}
\begin{figure}[h!]
    \centering
    \includegraphics[width=8.5cm, height=4cm]{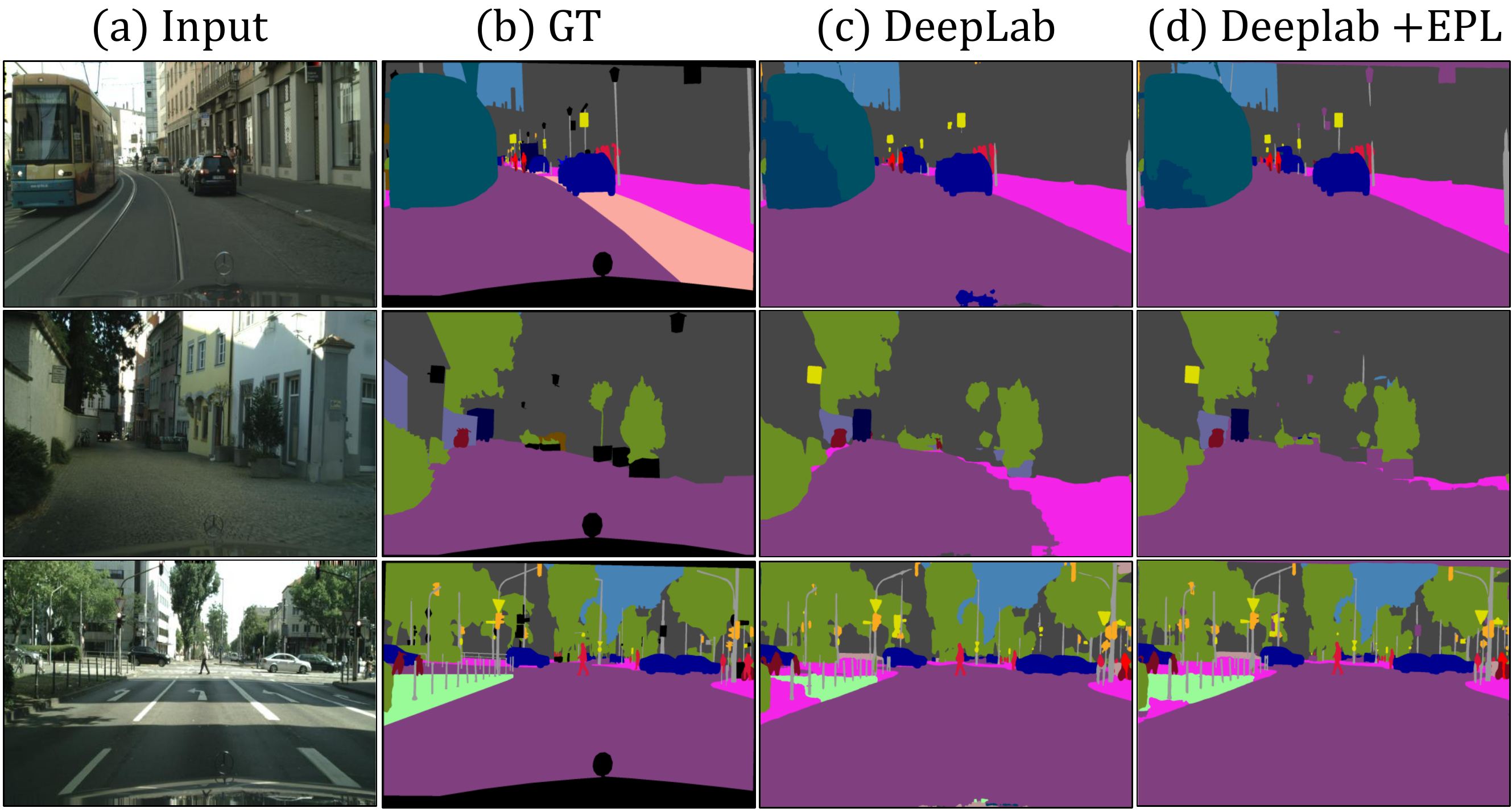}
    \caption{Qualitative comparison for segmentation results on Cityscapes validation set.}
    \label{fig:appendix_city_deep}
    \vspace{-10pt}
\end{figure}
\section{Additional semantic segmentation results}
In this part, we present more visual results of semantic segmentation experiments. We specifically choose images with challenging scenes, such as inter/intra-class cases, and images with misleading backgrounds. With demonstrated segmentation performance from \refFig{fig:appendix_pascal_psa} to \refFig{fig:appendix_city_deep}.